\documentclass[10pt,twocolumn,letterpaper,table]{article}

\usepackage{cvpr}
\usepackage{times}
\usepackage{epsfig}
\usepackage{graphicx}
\usepackage{amsmath}
\usepackage{amssymb}
\usepackage{bbm, bm}

\usepackage{tabularx}
\usepackage[export]{adjustbox}
\usepackage{array}
\usepackage{dsfont}
\usepackage{caption}
\usepackage{subcaption}
\usepackage{multirow}
\usepackage{transparent}

\usepackage{soul}
\usepackage[percent]{overpic}
\usepackage{adjustbox}
\usepackage{enumitem}
\usepackage[scientific-notation=true]{siunitx}

\usepackage[pagebackref=true,breaklinks=true,letterpaper=true,colorlinks,bookmarks=false]{hyperref}

\def\cvprPaperID{57}

\cvprfinalcopy
\ifcvprfinal\fi

\newcolumntype{L}[1]{>{\centering\arraybackslash}p{#1}}
\newcolumntype{Y}{>{\centering\arraybackslash}X}

\begin{document}

\title{Learning Action Maps of Large Environments via First-Person Vision}
\author{Nicholas Rhinehart, Kris M. Kitani\\
The Robotics Institute \\
Carnegie Mellon University\\
{\tt\small \{nrhineha, kkitani\}@cs.cmu.edu}
}

\maketitle

\newcommand{\nshscene}{Office Flr.\ A}
\newcommand{\slabscene}{Office Flr.\ B}
\newcommand{\gatesscene}{Office Flr.\ C}
\newcommand{\smithscene}{Office Flr.\ D}
\newcommand{\homescene}{Home A}

\newcommand{\nshsceneShort}{Of. Flr.\ A}
\newcommand{\slabsceneShort}{Of. Flr.\ B}
\newcommand{\gatessceneShort}{Of. Flr.\ C}
\newcommand{\smithsceneShort}{Of. Flr.\ D}
\newcommand{\homesceneShort}{Home A}

\newcommand{\objectness}{object detection}
\newcommand{\placeness}{scene classification}
\newcommand{\sitting}{\texttt{sitting}}
\newcommand{\typing}{\texttt{typing}}
\newif\ifbuildimage
\newif\ifbuildnewimage
\newif\ifbuildtables
\buildtablestrue
\buildnewimagetrue

\begin{abstract}
When people observe and interact with physical spaces, they are able to associate functionality to regions in the environment. Our goal is to automate dense functional understanding of large spaces by leveraging sparse activity demonstrations recorded from an ego-centric viewpoint. The method we describe enables functionality estimation in large scenes where people have behaved, as well as novel scenes where no behaviors are observed. Our method learns and predicts ``Action Maps", which encode the ability for a user to perform activities at various locations. With the usage of an egocentric camera to observe human activities, our method scales with the size of the scene without the need for mounting multiple static surveillance cameras and is well-suited to the task of observing activities up-close. We demonstrate that by capturing appearance-based attributes of the environment and associating these attributes with activity demonstrations, our proposed mathematical framework allows for the prediction of Action Maps in new environments. Additionally, we offer a preliminary glance of the applicability of Action Maps by demonstrating a proof-of-concept application in which they are used in concert with activity detections to perform localization.
\end{abstract}

\begin{figure}[t]
\begin{center}
\fcolorbox{black}{black}{\includegraphics[width=\linewidth]{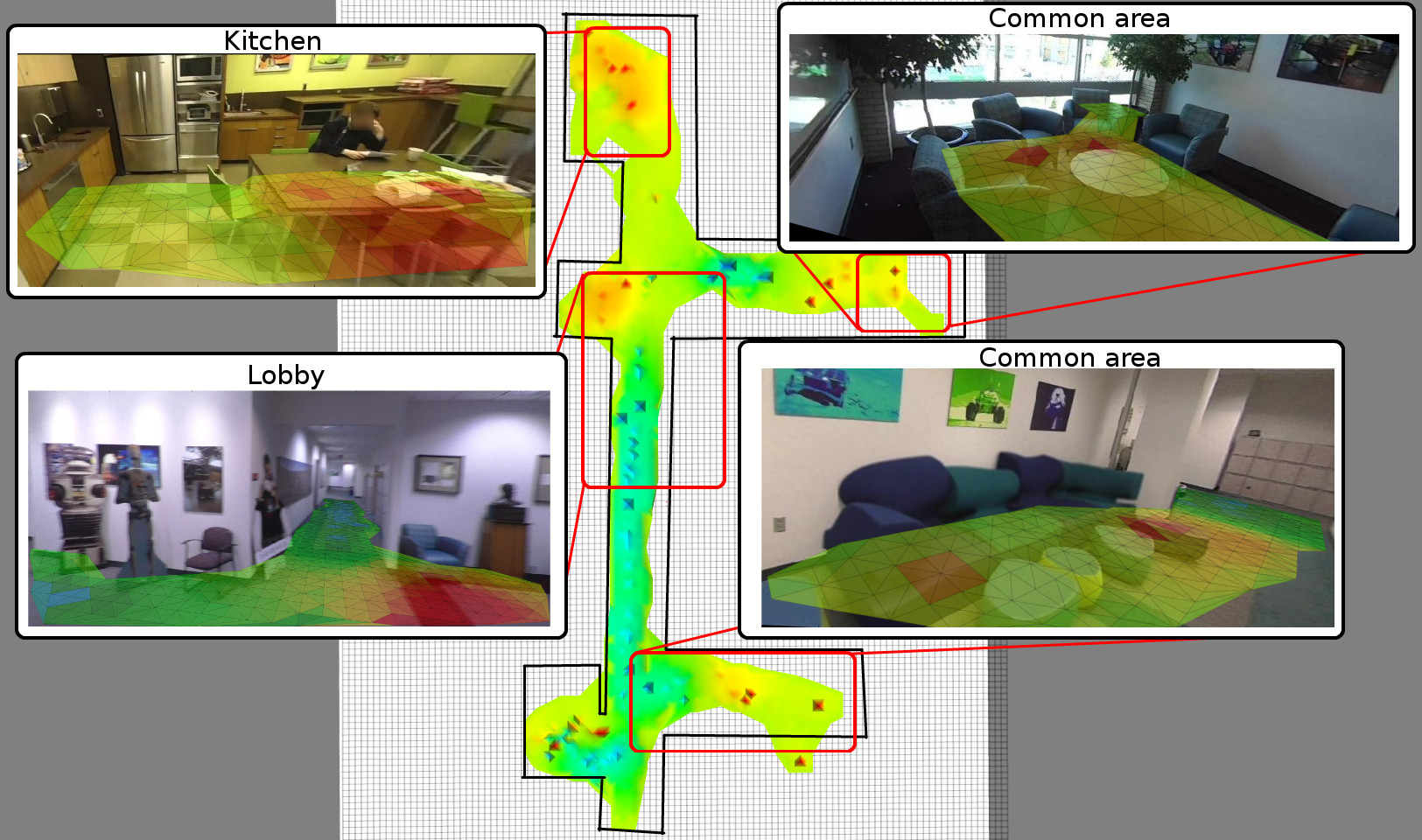}}\\
\fcolorbox{black}{black}{\includegraphics[width=\linewidth]{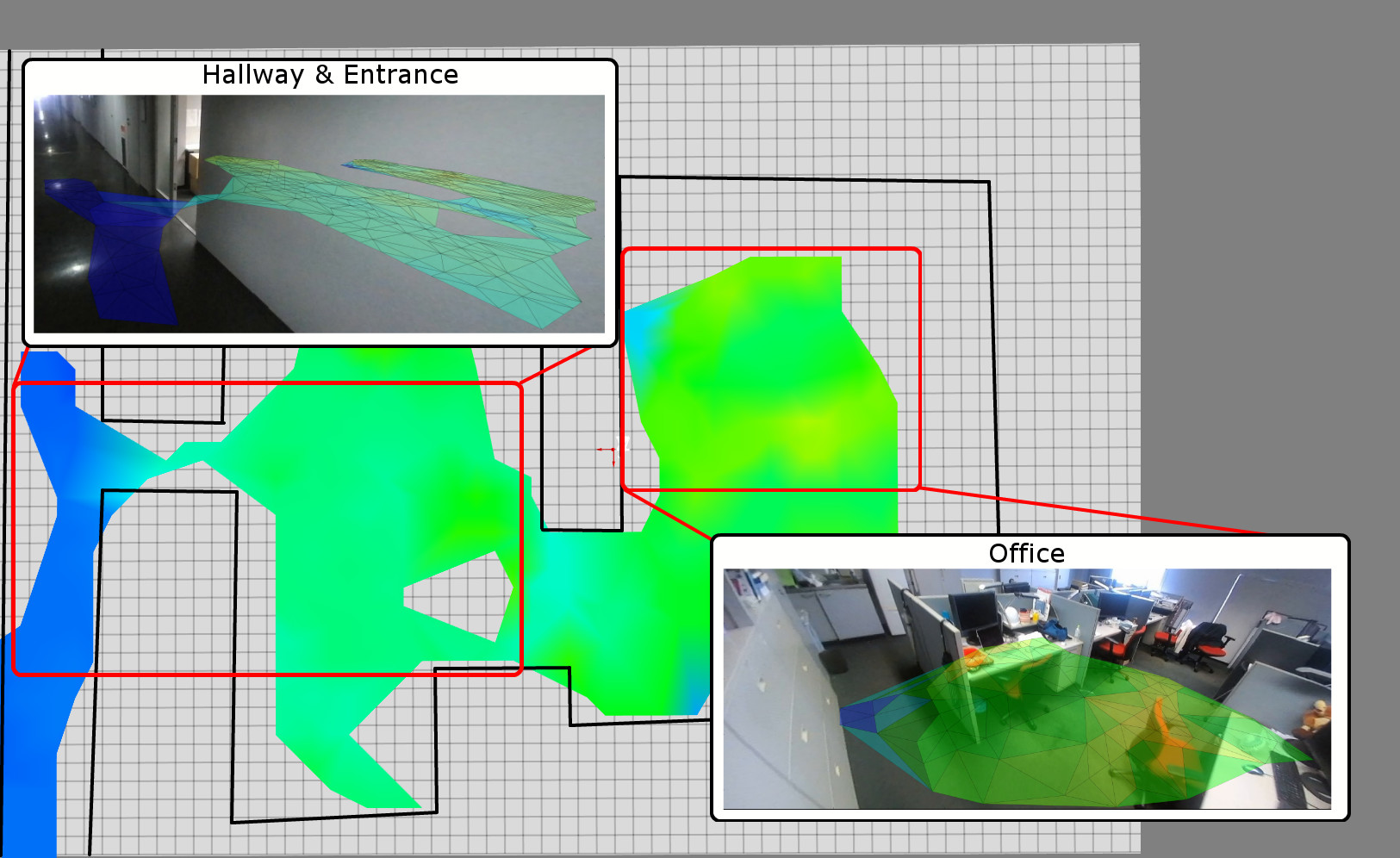}}
\fcolorbox{white}{white}{\includegraphics[width=\linewidth]{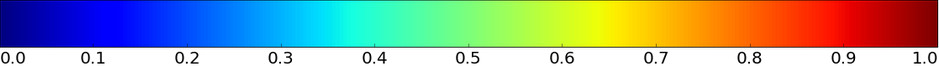}}
\end{center}
\vspace{-20pt}
  \caption{\small Action Map prediction for the \texttt{sit} activity by using our method to combine appearance data and activity observations. Activity and appearance information from the top scene in combination with \textit{only} appearance information (no activity observations) from the bottom scene is used to model the relationship between activities, scene information, and object information to make predictions for both scenes. Areas in the scenes where a person can sit are estimated by our method, such as the chairs and couches in both views.}
\label{fig:teaser}
\end{figure}

\section{Introduction}

The goal of this work is to endow intelligent systems with the ability to understand the functional attributes of their environment. Such functional understanding of spaces is a crucial component of holistic understanding and decision making by any agent, human or robotic. Functional understanding of a scene can range from the immediate environment to the distant. For example, at the scale of a single room,  a person can perceive the arrangement of tables, chairs, and  computers in an office environment, and reason that they could sit down and type at the computer.  People can also reason about the functionality about nearby rooms, for example, the presence of a kitchen  down the hall from the office is useful functional and spatial information for when the person decides to prepare a meal. The goal of this work is to learn a computational model of the functionality of large environments, called \textit{Action Maps} (AMs), by observing human interactions and the visual context of those action within a large environment.

There has been significant work in the area of automating the functional understanding of an environment, though much has focused on single scenes \cite{GuptaSatkinEfrosHebert_CVPR11, Fouhey12, koppula2013learning, jiang2013hallucinated, delaitre2012scene, gall2011functional}. In this work, we aim to extend automated functional understanding to very large spaces (\eg, an entire office building or home). This presents two key technical challenges:
\begin{itemize}
  \setlength\itemsep{2pt}
\item How can we capture observations of activity across large environments?
\item How can we generalize functional understanding to handle the inevitable data sparsity of less explored or new areas?
\end{itemize}

In order to address the first challenge of observing activity across large environments, we take a departure from the fixed surveillance camera paradigm, and propose an approach that uses a first-person point-of-view camera. By virtue of its placement, its view of the wearer's interactions with the environment is usually unobstructed by the wearer's body and other elements in the scene. An egocentric camera is portable across multiple rooms, whereas fixed cameras are not. An egocentric camera allows for the observation of hand-based activities, such typing or opening doors, as well as the observation of some ego-motion based activities, such as sitting down or standing. The first-person paradigm is well suited for large-scale sensing and allows observation of interactions with many environments.

Although we can capture a large number of observations of activity across large environments with wearable cameras, it is still not practical to wait to observe all possible actions in all possible locations. This leads to the second technical challenge of generalizing functional understanding from a \textit{sparse} set of action observations, which requires generalization to new locations. Our method generalizes by using another source of visual observation -- which we call \textit{side-information} -- that encodes per-location cues relevant to activities. In particular, we propose to extract visual side-information using scene classification \cite{zhou2014learning} and object detection \cite{girshick14CVPR} techniques. With this information, our method learns to model the relationship between actions, scenes, and objects. In a scene with no actions, we use scene and object information, coupled with actions in a separate scene, to infer possible actions. We propose to solve the problem of generalizing functional understanding (\ie, generating dense AMs) by formulating the problem as matrix completion. Our method constructs a matrix where each row represents a location and each column represents an action type (\eg, read, sit, type, write, open, wash). The goal of matrix completion is to use the observed entries to fill the missing entries. In this work, we make use of Regularized Weighted Non-Negative Matrix Factorization (RWNMF) \cite{gu2010collaborative}, allowing us to elegantly leverage \textit{side-information} to model the relationship between activities, scenes, and objects, and predict missing activity affordances.

\newlength{\figtextoff}
\setlength{\figtextoff}{6mm}
\newlength{\projactmapfigwidth}
\setlength{\projactmapfigwidth}{.23\textwidth}
\ifbuildnewimage
\begin{figure}[ht]
\centering
\footnotesize
\begin{subfigure}[c]{\projactmapfigwidth}
\centering
\begin{overpic}[height=1in,tics=10]{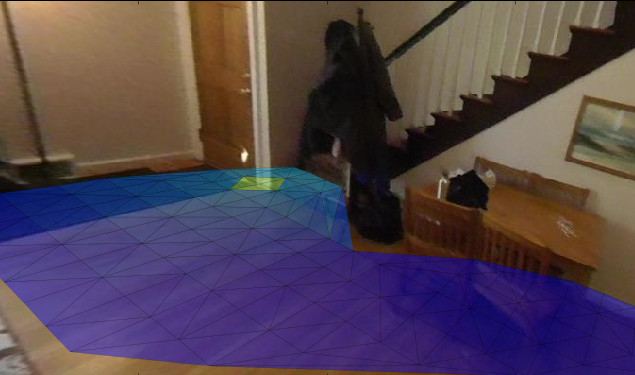} \put(0,52){\transparent{0.6}\colorbox{black}{ \transparent{1.0}\color{white}{\scriptsize \textbf{Estimated \texttt{opendoor} Action Map}}}}
\end{overpic}
\end{subfigure}\hfill
\begin{subfigure}[c]{\projactmapfigwidth}
\centering
\begin{overpic}[height=1in,tics=10]{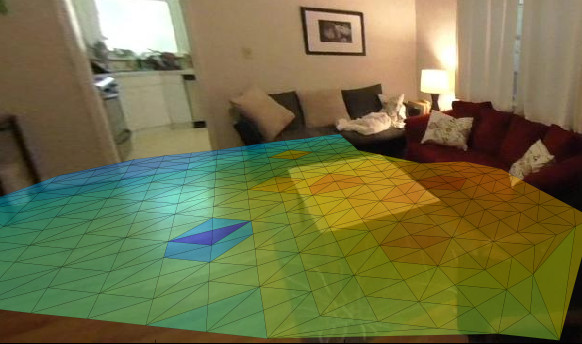} 
\put(0,52){\transparent{0.6}\colorbox{black}{\transparent{1.0}\color{white}{\scriptsize \textbf{Estimated \texttt{sit} Action Map}}}}
\end{overpic}
\end{subfigure}
\vspace{4pt}
\begin{subfigure}[c]{\projactmapfigwidth}
\centering
\begin{overpic}[height=1in,tics=10]{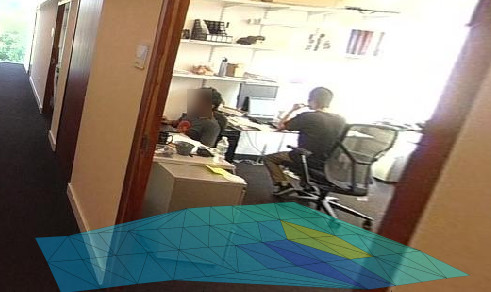} 
\put(0,53){\transparent{0.6}\colorbox{black}{\transparent{1.0}\color{white}{\scriptsize \textbf{Estimated \texttt{typing} Action Map}}}}
\end{overpic}
\end{subfigure}\hfill
\begin{subfigure}[c]{.228\textwidth}
\centering
\begin{overpic}[height=1in,tics=10]{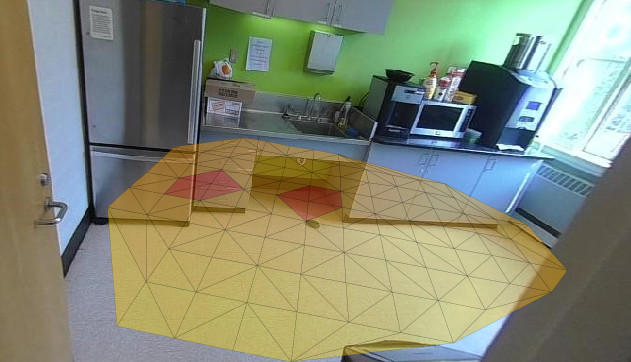} 
\put(0,51){\transparent{0.6}\colorbox{black}{\transparent{1.0}\color{white}{\scriptsize \textbf{Estimated \texttt{wash} Action Map}}}}
\end{overpic}
\end{subfigure}
\vspace{-7pt}
\caption{\small Projected Action Map examples learned by our method. With global estimates of large Action Maps produced by our method, we use localized images within the scene to show visualizations of the Action Maps by projecting them to the images.}
\vspace{-12pt}
\label{fig:project-action-map-examples}
\end{figure}
\fi
\subsection{Contributions}

To the best of our knowledge, this is the first work to generate \textit{Action Maps}, such as those in Figures~\ref{fig:teaser}~and~\ref{fig:project-action-map-examples}, over large spaces using a wearable camera. The first-person vision paradigm is an essential tool for this problem, as it can capture a wide range of visual information across a large environment. Our approach unifies scene functionality information via a regularized matrix completion framework that appropriately addresses the issue of sparse observations and provides a vehicle to leverage visual side information. 

We demonstrate the efficacy of our proposed approach on five different multi-room scenes: one home and four office environments. Our experiments in real large-scale environments show how first-person sensing can be used to efficiently observe human activity along with visual side-information across large spaces. \textbf{1)} We show that our method can be used to model visual information from both single and multiple scenes simultaneously, and makes efficient use of all available activity information. \textbf{2)} We show that our method's power increases as the set of performed activity increases. \textbf{3)} Furthermore, we demonstrate how our proposed matrix factorization framework can be used to leverage sparse observations of human actions along with visual side-information to perform functionality estimation of large \textit{novel} scenes in which \textit{no activities} have been demonstrated. We compare our proposed method against natural baselines such as object-detection-based Action Maps and scene classification, and show that our approach outperforms them in nearly all of our experiments. \textbf{4)} Additionally, as a proof-of-concept application of the rich information in an Action Map, we present an application of our Action Maps as priors for localization.

\subsection{Background}

Human actions are deeply connected to the scene. Scene context (\eg, a chair or common room) can be a strong indicator of actions (\eg, sitting). Likewise, observing an action like sitting, is a strong indicator that there must be a sittable surface in the scene. In the context of time lapse video, Fouhey \etal \cite{Fouhey12} used detection of sitting, standing, and walking actions to obtain better estimates of 3D geometry for a single densely explored room. Gupta \etal \cite{GuptaSatkinEfrosHebert_CVPR11} addressed the inverse problem of inferring actions from estimated 3D scene geometry using a single image of a room. Their approach synthetically inserted skeleton models into the 3D scene to reason about possible functional attributes of the scene. Delaitre \etal \cite{delaitre2012scene} also used time lapse video of human actions to learn the functional attributes of objects in a single scene. The work of Savva \etal  \cite{savva2014scenegrok} obtains a dense 3D representation of small workspace (\eg desk and chair space) and learns the functional attributes of the scene by observing human interactions. Similar to previous work, this work seeks to understand the functionality of scenes. However, limitations of previous work include the reduced size of the physical space and the presumed density of interactions. In contrast, our approach attempts to infers the dense functionality over an entire building (\eg, office floor or house), and reasons about multiple large scenes simultaneously by modeling the relationship between scene information, object information, and sparse activities.

Another flavor of approaches reason in the joint space of activities and objects. In Moore \etal \cite{moore1999exploiting}, human actions are recognized by using information about objects in the scene. Gall \etal \cite{gall2011functional} uses human interaction information to perform unsupervised categorization of objects. Other approaches have capitalized on the interplay between actions and objects:  Gupta \etal \cite{gupta2008context} demonstrate an approach to use object information for pose detection, and Yao \etal \cite{yao2010modeling} jointly model objects and poses to perform recognition of both objects and actions. The approach of \cite{peursum2005combining} performs object recognition by observing human activities, and notes an important idea that our approach also uses: whereas object information may sometimes be too small in detail, human activities usually are not. We capitalize on this observation close-up observation capability of an egocentric camera.

The egocentric paradigm is an excellent method for understanding human activities at close range \cite{De_la_Torre_Frade_2009_6394, fathi2011understanding, pirsiavash2012detecting, li2015delving}. Our work builds on such egocentric action recognition techniques by associating actions with physical locations in a single holistic framework. By bringing together ideas from single image functional scene understanding, object functionality understanding and egocentric action analysis, we propose a computational model that enables cross-building level functional understanding of scenes.

\ifbuildimage
\begin{figure}[t]
\begin{center}
\includegraphics[width=.6\linewidth]{images/diagram_q90.jpg}
\caption{Flowchart detailing the high-level components of our approach. Egocentric video is used to localize a user and detect their actions, which are integrated into a representation relating the physical space in a scene to the affordances of actions (`Action Map').} \label{fig:flowchart}
\end{center}
\end{figure}
\fi

\section{Constructing Action Maps} \label{sec:action_maps}

Our goal is to build \textit{Action Map}s that associate possible actions for every spatial location on a map over a large environment. We decompose the process into three steps. We first build a physical map of the environment by using egocentric videos to obtain a 3D reconstruction of the scene using structure from motion. Second, we use a collection of recorded human activity videos recorded with an egocentric camera to detect and spatially localize actions. This collection of videos is also used to learn the visual context of actions (\ie, scene appearance and object detections) which is later used as a source of side information for modeling and inference. Third, we aggregate the localized action detection and visual context data using a matrix completion framework to generate the final Action Map. The focus of our method is the third step, which we describe next. We mention how we obtain the visual context in Section~\ref{sec:side-information-graph}, and describe the first two steps in detail in Section~\ref{sec:preprocessing}.

\ifbuildnewimage
\begin{figure*}[!t]
\centering
\footnotesize
\begin{subfigure}[c]{.45\textwidth}
\includegraphics[width=\textwidth]{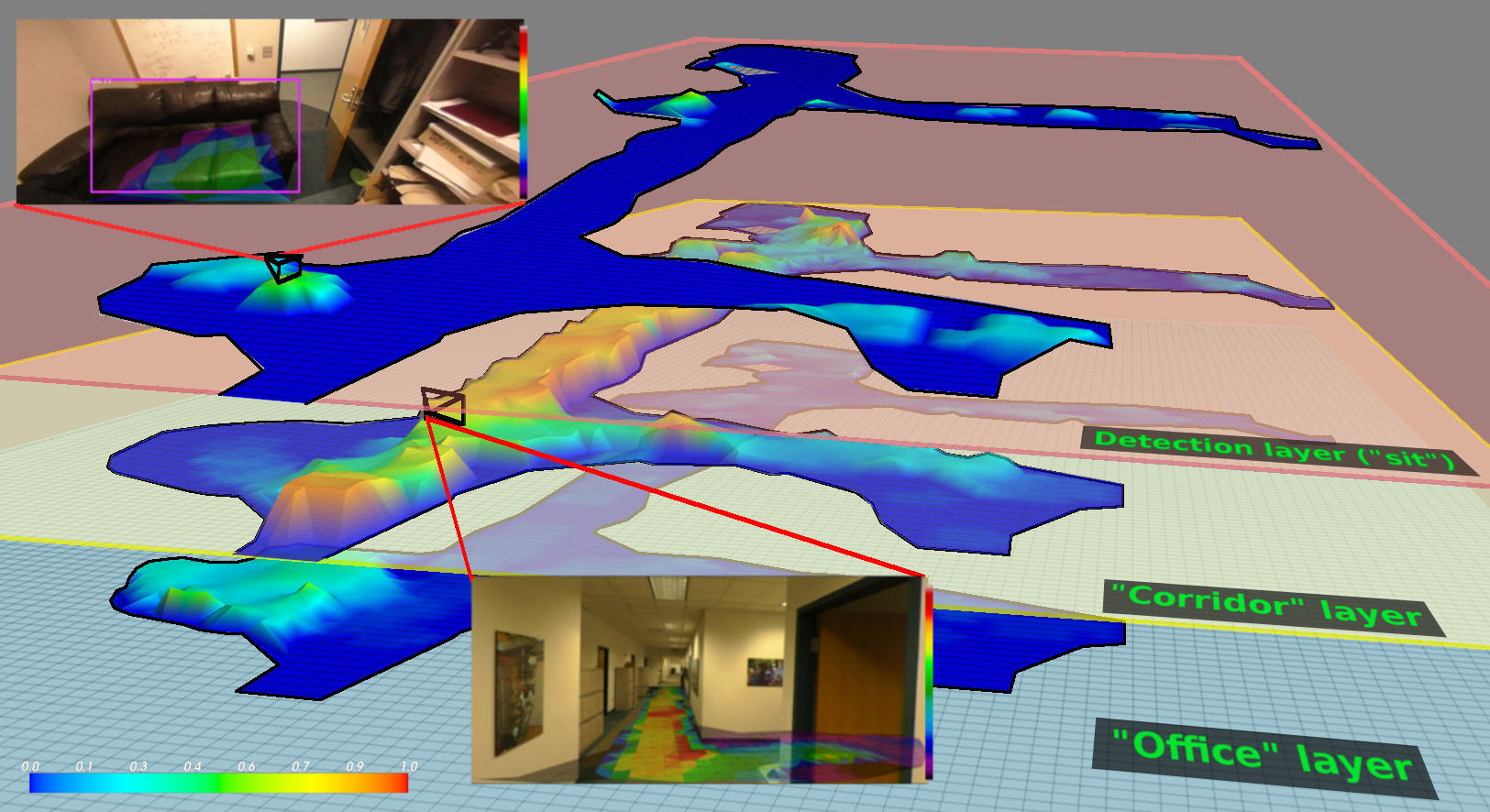}
\caption{\small \nshscene~Features}
\end{subfigure}
\begin{subfigure}[c]{.45\textwidth}
\includegraphics[width=\textwidth]{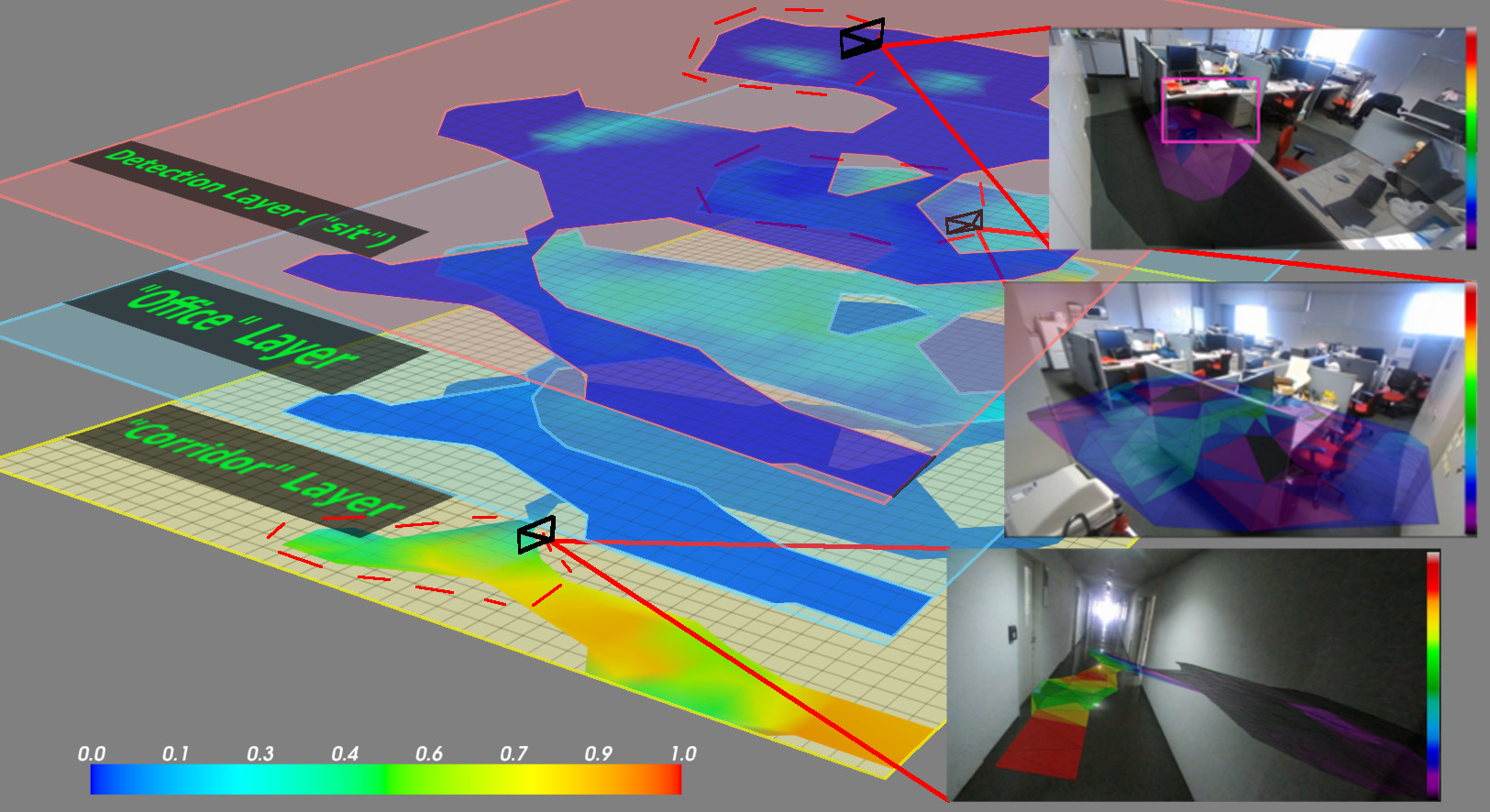}
\caption{\small \slabscene~Features}
\end{subfigure}
\vspace{-7pt}
\caption{\small Several \nshscene~and \slabscene~Features. The ``office" and ``corridor" layers correspond to the features from the \placeness~CNN, and the ``sit" layer corresponds to the \objectness~CNN features aggregated across all \texttt{sit}-able objects, which is also one of the baselines as described in Section~\ref{sec:preprocessing}. This figure demonstrates our idea that object information and scene information can be used to relate scenes to each other. This relationship is the basis for transferring and sharing activity functionality between scenes. Heatmaps from several layers are shown projected into localized images from the scene. Note that the ``office" portion of \nshscene~also contains sittable regions, and that the much larger ``office" area in \slabscene~contains a select few sittable regions. The corridors in both scenes are described well by the features, and these areas strongly correlate with an \textit{an absence} of functionality, as scene in Figure~\ref{fig:teaser}.} \label{fig:nsh-feature-examples}
\end{figure*}
\fi

\subsection{Action Map Prediction as Matrix Factorization} \label{sec:matrix_factorization}

We now describe our method for integrating the sparse set of localized actions and visual side-information to generate a dense Action Map (AM) using regularized matrix completion. Our goal is to recover an AM in matrix form $\mathbf{R} \in \mathbb{R}_{+}^{M\times A}$, where $M$ is the number of locations on the discretized ground plane and $A$ is the number of possible actions. Each row of the AM matrix $\mathbf{R}$ contains the action scores $\mathbf{r}_m$, where $m$ is a location index, and each entry $r_{ma}$ describes the extent to which an activity $a$ can be performed at location $m$. To complete the missing entries of $\mathbf{R}$, we design a similarity metric for our side-information, enabling the method to model the relationship between activities, scenes, and objects.

We impose structure on the rows and columns of the AM matrix by computing similarity scores with the side-information. Examples of this side information are shown in Figure~\ref{fig:nsh-feature-examples}, where two features from scene classification, plus one feature from object detection are shown in the same physical space as the AM. Figure~\ref{fig:nsh-feature-examples} serves to further motivate the idea of exploiting scene and object information between two different scenes to relate the functionality of the scenes. We define three kernel functions based on scene appearance, object detections and spatial continuity. This structure is integrated as regularization in the RWNMF objective function (Equation~\ref{eq:wnmf_reg}).

\subsubsection{Integrating Side-Information} \label{sec:side-information-graph}

To integrate side-information into our formulation, we build two weighted graphs that describe the cross-location (row) similarities, and cross-action (column) similarities. 
We are primarily interested in the cross-location similarities, and discuss how we handle the cross-action similarities in Section~\ref{sec:obj-func-min}. To build the cross-location graph, we aggregate the spatial proximity, scene-classification, and object detection information as a linear combination of kernel-based similarities, as shown in Equation~\ref{eq:kernel-combo}. 

For every location $a$ in the AM, we compute the \placeness~score $\mathbf{p}_{a} = \left[p_{1a} \dots p_{Ca}\right]$ for each image as the average of the $C$-dimensional outputs from the Places-CNN of images within a small radius.

We use Structure-from-Motion (SFM) keypoints inside each detection to estimate the back-projected 3D location of the detected object in the environment by taking the mean of their 3D locations, which are then projected to the ground plane to form a set $\mathcal{D}_f$ for each object category $f \in \left[1 \dots F \right]$. The SFM reconstruction is also used to localize images and described further in Section~\ref{sec:preprocessing}. We calculate the \objectness~scores $\mathbf{o}_a = \left[o_{1a} \dots o_{Fa}\right]$ for each location $a$ as the max score of object detection of the nearby back-projected object detections $d \in \mathcal{D}_f$ within a $r=\sqrt{2}$ grid-cell radius, exponentially weighted by its distance along the floor from the object $z_d$:  
\begin{align*}
o_{fa} = &\max_{d \in \mathcal{D}_f}\frac{1}{\sqrt{2r^2\pi}}\exp{\frac{-z_d^2}{2r^2}}.
\end{align*}

We wish to enforce similarity of activities between nearby locations, as well as between locations that have similar object detections and scene classification description. Between any two locations $a$, $b$, and given associated \placeness~scores $\mathbf{p}_{a}, \mathbf{p}_{b}$, \objectness~scores~$\mathbf{o}_{a}, \mathbf{o}_{b}$, and 2D grid locations $\mathbf{x}_a, \mathbf{x}_b$ the kernel is of the form:
\begin{align}\label{eq:kernel-combo}
\begin{split}
k(a, b) = (1 - \alpha) & k_\text{s}(\mathbf{x}_a, \mathbf{x}_b) + \\
\frac{\alpha}{2} & k_\textit{p}(\mathbf{p}_a, \mathbf{p}_{b}) + \frac{\alpha}{2}k_\textit{o}(\mathbf{o}_a, \mathbf{o}'_{b}),
\end{split}
\end{align}
where $k_s$ is an RBF kernel between the spatial coordinates of each location, $k_p$ and $k_o$ as $\chi^2$ kernels on \placeness~scores and \objectness~scores, and $k_o$ has 0 similarity between locations with no object score. 

Thus, there is a tradeoff between the $k_s$, $k_p$ and $k_o$ kernels, controlled by $\alpha$. When $\alpha = 0$, only spatial smoothness is considered, and when $\alpha = 1$, only \placeness~and \objectness~terms are considered, ignoring spatial smoothness. When a location in one scene is compared to a location in a new scene or the same scene, $k(\cdot,\cdot)$ returns higher scores for locations with similar objects and places, and as shown Section~\ref{sec:obj-func-min}, places more regularization constraint on the objective function, rewarding solutions that predict similar functionalities for \textit{both} locations.

\subsection{Completing the Action Map Matrix} \label{sec:obj-func-min}

To build our model, we seek to minimize the RWNMF objective function in Equation~\ref{eq:wnmf_reg}:

\begin{equation}
\begin{split}\label{eq:wnmf_reg}
J(\mathbf{U}, \mathbf{V}) = \left \| \mathbf{W} \circ (\mathbf{R} - \mathbf{UV}^T) \right\|_F^2 & + \frac{\lambda}{2}\sum_{i,j}^M\left\|\mathbf{u}_i - \mathbf{u}_j\right\|\mathbf{K}^U_{ij}  \\
              &   + \frac{\mu}{2}\sum_{i,j}^A \left\|\mathbf{v}_i - \mathbf{v}_j \right\|\mathbf{K}^{V}_{ij}
\end{split}
\end{equation}

where $\mathbf{U} \in \mathbb{R}_{+}^{M\times D}$, $\mathbf{V} \in \mathbb{R}_{+}^{A\times D}$, together form the decomposition, $\mathbf{W} \in \mathbb{R}_{+}^{M \times A}$ is the weight matrix with $0$s for unexplored locations, and $\mathbf{K}^U$ the kernel Gram matrix of the side information defined by its elements: $\mathbf{K}^U_{ij} = k(i, j)$. The squared-loss term penalizes decompositions with values different from the observed values in $\mathbf{R}$. The term involving $\mathbf{K}^U$ penalizes decompositions in which highly similar locations have different decompositions in the rows ($\mathbf{u}_i^T$) of $\mathbf{U}$. Roughly, locations with high similarity in scene appearance, object presence, or position impose penalty on the resulting decomposition for predicting different affordance values in the AM. The term involving $\mathbf{K}_V$ corresponds to the cross-action smoothing, which we take as the identity matrix, enforcing no penalty for differences across per-location action labels.

To minimize the objective function, we use the regularized multiplicative update rules following \cite{gu2010collaborative}.
Multiplicative update schemes for NMF are generally constructed such that their iterative application yields a non-increasing update to the objective function; \cite{gu2010collaborative} showed that these update rules yield non-increasing updates to the objective function. Thus, after enough iterations, a local minima in the objective function is found, and the resulting decomposition and its predictions are returned.

Values in $\mathbf{W}$ are set to counteract class imbalance. The number of observed values for each activity is computed as $n_c$, and assigned to each nonempty location $i$'s corresponding entry as $w_{ic} = 1 /n_c$, and the zeros from observed cameras associated with no activities as $w = 1 / n_z$.

\section{Experiments}\label{sec:experiments}
\ifbuildtables
\begin{table}[t]
\centering
\footnotesize
\begin{tabular}{|c|c|c|c|c|c|c|c|c|c|}
\hline
        & \# GT locs. & \# Actions  & Length & $r_e$ & $r_a$\\
        \hline
\nshscene &         40        &         90   & 53.3 min. &  0.59 &  0.03\\
\hline
\smithscene &       15          &        44    & 32.8 min. & 0.23 & 0.03\\
\hline
\gatesscene &       44          &        14  & 12.2 min. & 0.16 &  0.01 \\
\hline
\slabscene &         50        &        13  & 3.3 min. & 0.67 &  0.04\\
\hline
\homescene &         15       &     17     & 13.4 min.  & 0.75 & 0.04 \\
\hline
\end{tabular}
\vspace{-5pt}
   \caption{\small Scene stats. The number of GT locations is the number of distinct places a specific activity can be performed. The number of activity demonstrations is the total number of demonstrations collected in each environment.
    $r_e = \frac{\# \text{cells explored}}{\# \text{total cells}}$, $r_a = \frac{\# \text{cells with non-empty actions}}{\# \text{total cells}}$.}
\label{tab:scene_stats}
\end{table}
\fi

\begin{table}[t]
\setlength{\tabcolsep}{3pt}
\centering
\footnotesize
\begin{tabular}{|l||c|l|c|l|}%
\hline
                      & W. Max F1 & W. Mean F1 & Max F1 & Mean F1 \\
\hline
\hline
\nshsceneShort~S sng & 0.73 & 0.72 $\pm$ 0.01 & 0.44 & 0.43 $\pm$ 0.02 \\ 
\nshsceneShort~SOP$_D$ sng & 0.63 & 0.61 $\pm$ 0.01 & 0.34 & 0.32 $\pm$ 0.01\\
\nshsceneShort~SOP sng & 0.74 & 0.69 $\pm$ 0.04 & {\bf 0.56} & 0.5 $\pm$ 0.04  \\ %
\hline
\nshsceneShort~SOP$_D$ all & 0.75 & 0.71 $\pm$ 0.02 & 0.44 & 0.43 $\pm$ 0.01 \\
\nshsceneShort~SOP all & {\bf 0.76} & {\bf 0.73} $\pm$ 0.02 & 0.54 & {\bf 0.51} $\pm$ 0.02 \\%& 0.91 & 0.79 & 0.51 & 0.4 & 0.16 & 0.14 & 0.086 & 0.018 & 0.68 & 0.37 & 0.89 & 0.79

\hline \hline
\slabsceneShort~S sng &  0.56 & 0.55 $\pm$ 0.01 & 0.38 & 0.38 $\pm$ 0.01 \\
\slabsceneShort~SOP sng & 0.56 & 0.55 $\pm$ 0.01  & 0.44 & 0.38 $\pm$ 0.03 \\  %
\hline
\slabsceneShort~SOP$_D$ all & {\bf 0.58} & {\bf 0.56} $\pm$ 0.01 & 0.39 & 0.37 $\pm$ 0.03 \\%%

\slabsceneShort~SOP all & {\bf 0.58} & {\bf 0.56} $\pm$ 0.01 & {\bf 0.53} & {\bf 0.44} $\pm$ 0.04 \\%& 0.67 & 0.62 & 0.15 & 0.091 & 0.48 & 0.21 & 0.87 & 0.2 & 0.38 & 0.37 & 0.63 & 0.58
\hline  \hline
\gatessceneShort~S sng & 0.74 & {\bf 0.66} $\pm$ 0.1 & 0.48 & 0.42 $\pm$ 0.06 \\
\gatessceneShort~SOP$_D$ sng & 0.67 & 0.46 $\pm$ 0.08 & 0.41 & 0.29 $\pm$ 0.05 \\%%
\gatessceneShort~SOP sng & 0.68 & 0.53 $\pm$ 0.1 & 0.53 & 0.44 $\pm$ 0.06 \\%& 0.68 & 0.43 & 0.2 & 0.062 & 0.34 & 0.17 & 0.42 & 0.22 & 0.69 & 0.51 & 0.88 & 0.61 \\
\hline
\gatessceneShort~SOP$_D$ all & 0.67 & 0.55 $\pm$ 0.06 & 0.45 & 0.38 $\pm$ 0.03 \\
\gatessceneShort~SOP all & {\bf 0.77} & 0.58 $\pm$ 0.07 & {\bf 0.56} & {\bf 0.46} $\pm$ 0.04 \\%& 0.74 & 0.55 & 0.39 & 0.23 & 0.18 & 0.11 & 0.18 & 0.09 & 0.97 & 0.61 & 0.88 & 0.6
\hline \hline
\smithsceneShort~S sng & 0.68 & 0.57 $\pm$ 0.11 & 0.57 & 0.45 $\pm$ 0.12 \\
\smithsceneShort~SOP$_D$ sng & 0.56 & 0.49 $\pm$ 0.05 & 0.37 & 0.32  $\pm$ 0.04\\
\smithsceneShort~SOP sng & 0.69 & 0.55 $\pm$ 0.08 & 0.68 & 0.54 $\pm$ 0.07 \\%%
\hline
\smithsceneShort~SOP$_D$ all & 0.81 & 0.68 $\pm$ 0.07 & 0.59 & 0.46 $\pm$ 0.08 \\
\smithsceneShort~SOP all & {\bf 0.82} & {\bf 0.73} $\pm$ 0.08 & {\bf 0.77} & {\bf 0.61} $\pm$ 0.09 \\%& 0.89 & 0.65 & 0.6 & 0.12 & 0.57 & 0.21 & 0.87 & 0.34 & 0.88 & 0.48 & 0.79 & 0.47
\hline  \hline
\homesceneShort~S sng &  0.57 & 0.53 $\pm$ 0.04 & 0.35 & 0.34 $\pm$ 0.02 \\
\homesceneShort~SOP$_D$ sng & 0.5 & 0.48 $\pm$ 0.01 & 0.26 & 0.24 $\pm$ 0.02 \\
\homesceneShort~SOP sng & {\bf 0.62} & {\bf 0.6} $\pm$ 0.01 & 0.43 & {\bf 0.4} $\pm$ 0.02 \\%%
\hline
\homesceneShort~SOP$_D$ all & 0.52 & 0.49 $\pm$ 0.03 & 0.27 & 0.25 $\pm$ 0.02 \\ 
\homesceneShort~SOP all & {\bf 0.62} & 0.55 $\pm$ 0.03 & {\bf 0.45} & {\bf 0.4}$\pm$0.02 \\ %
\hline
\end{tabular}
\vspace{-3pt}
\caption{\small Prediction results by using the activity observations for each scene (``sng"), and, as separate results, by simultaneously fitting data from all scenes (``all"). By using observations from all scenes, the performance of our method on each scene improves over using each scene's observation data alone. Additionally, our method is able to integrate activity detections without much performance loss: a $_D$ suffix indicates activity detection predictions were used, otherwise, labelled activities were used. ``S" stands for spatial kernel only, and ``SOP" stands for ``Spatial+Object Detection+Scene Classification" kernels. The spatial kernel only is useful yet outperformed by the full model. Side information from multiple scenes generally improves the performance.} \label{tab:all_scenes_observations}
\end{table}

\ifbuildnewimage
\begin{figure*}[ht]
\centering
\footnotesize
\begin{subfigure}[c]{.29\textwidth}
\includegraphics[width=\textwidth]{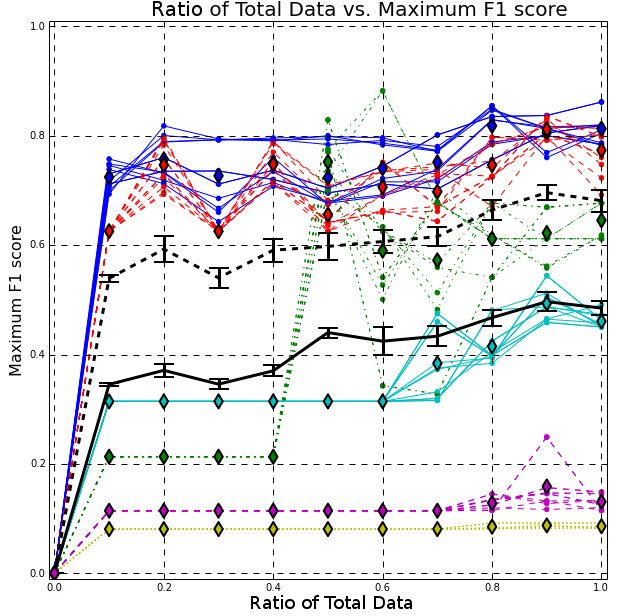}
\caption{\nshscene~Elapse}
\end{subfigure}
\begin{subfigure}[c]{.29\textwidth}
\includegraphics[width=\textwidth]{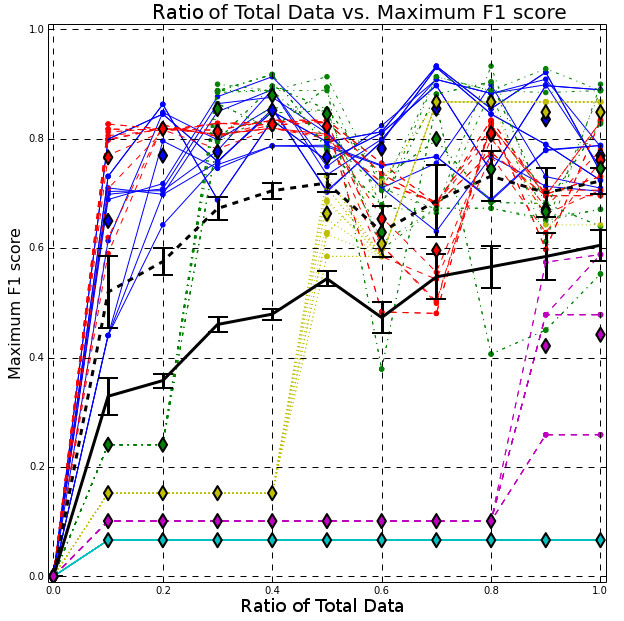}
\caption{\smithscene~Elapse}
\end{subfigure}
\begin{subfigure}[c]{.29\textwidth}
\includegraphics[width=\textwidth]{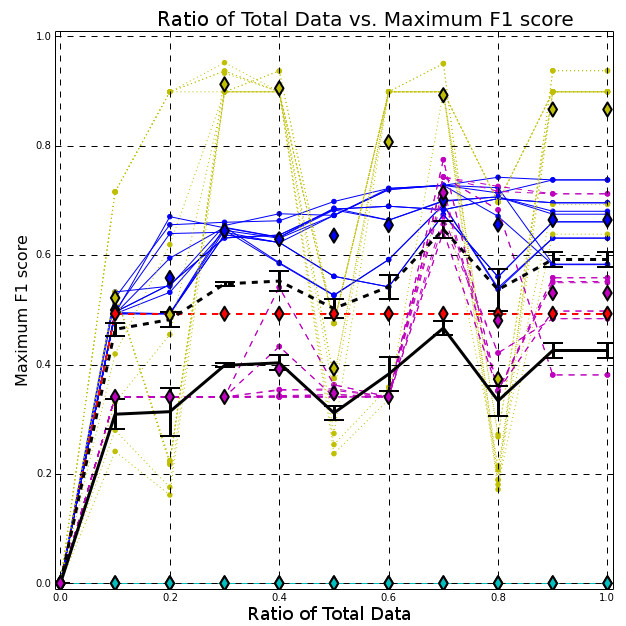}
\caption{\homescene~Elapse}
\end{subfigure}
\begin{subfigure}[t]{.11\textwidth}
\includegraphics[width=\textwidth]{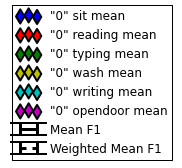}
\end{subfigure}
\vspace{-7pt}
\caption{\small Performance improves a function of available data. For each parameter setting, we show the $F1$ scores for each
activity label, as well as the mean and weighted mean of the $F1$ scores across all parameter settings and 
activity labels. Some variations in performance are observed as new activities are introduced, as the correlations between an established activities and newly introduced activities are initially sparse. As more data is collected, erroneous correlations are unlearnt, and correct ones are reinforced.} \label{fig:elapse-graphs}
\end{figure*}
\fi

\begin{table*}[t]
\vspace{-5pt}
\centering
\footnotesize
\setlength{\tabcolsep}{4pt}
\begin{tabular}{|c|l||c|c|c|c||c|l||c|c|c|}%
\hline
           &           & W. Max F1 & W. Mean F1 & Max F1 & Mean F1 &           & W. Max F1 & W. Mean F1 & Max F1 & Mean F1 \\
\hline
\hline
\multirow{ 6}{*}{\slabscene} & RFC & 0.38 & 0.38 & {\bf 0.62 } & {\bf 0.62} & \multirow{ 6}{*}{\smithscene}  & 0.27 & 0.27 & 0.41 & {\bf 0.41} \\
&Det. &  0.59& 0.59 & 0.33 & 0.33 & &  0.44 & 0.44 & 0.28 & 0.28 \\
& NMF & 0.35 & 0.35 & 0.24 & 0.24 & & 0.65 & 0.65 & 0.40 & 0.40 \\ %
\cline{2-6} \cline{8-11}
&{\bf SO} & 0.69 & 0.67 $\pm$ 0.02 & 0.44 & 0.42 $\pm$ 0.01 &  & 0.65 & 0.51 $\pm$ 0.12 & 0.46 & 0.36 $\pm$ 0.09\\%%
&{\bf SP} &  {\bf 0.74} & {\bf 0.69} $\pm$ 0.02 & 0.46 & 0.43 $\pm$ 0.02 & & {\bf 0.68} & {\bf 0.55} $\pm$ 0.12 & {\bf 0.51} & 0.38 $\pm$ 0.09\\%%
&{\bf SOP} & 0.57 & 0.54 $\pm$ 0.03 & 0.28 & 0.26 $\pm$ 0.02 & & 0.42 & 0.36 $\pm$ 0.02 & 0.28 & 0.25 $\pm$ 0.01  \\%%
\hline
\hline
\multirow{ 6}{*}{\gatesscene}&RFC & 0.24 & 0.24 & 0.37 & 0.37 & \multirow{ 6}{*}{\homescene} & 0.28 & 0.28 & 0.35 & 0.35\\
&Det. &  0.54 & 0.54 & 0.31 & 0.31  &  & 0.53 & 0.53 & 0.25 & 0.25 \\
&NMF & 0.39 & 0.39 & 0.27 & 0.27 &  & 0.43 & 0.43 & 0.25 & 0.25 \\ %
\cline{2-6} \cline{8-11}
&{\bf SO} & 0.67 & 0.55 $\pm$ 0.1 & 0.47 & 0.39 $\pm$ 0.07 & & 0.59 & 0.51 $\pm$ 0.07 & 0.41 & 0.33\\ %
&{\bf SP} & 0.61 & 0.56 $\pm$ 0.08 & 0.47 & 0.39 $\pm$ 0.06 &  & {\bf 0.61} & {\bf 0.58} $\pm$ 0.01 & {\bf 0.45} & {\bf 0.42} $\pm$ 0.03\\%%
&{\bf SOP} & {\bf 0.74} & {\bf 0.63} $\pm$ 0.05 & {\bf 0.64} & {\bf 0.54} $\pm$ 0.05 & & 0.54 & 0.45 $\pm$ 0.03 & 0.3 & 0.26 $\pm$ 0.01\\%& 0.66 & 0.39 & 0.39 & 0.22 & 0.78 & 0.31 & 0.21 & 0.065 & 0.92 & 0.37 & 0.85 &0.38'
\hline

\end{tabular}
\vspace{-4pt}
\caption{\small Performance of our algorithm by using activity observations from \nshscene~to make predictions in novel scenes. 
Each baseline method is run with a single parameter setting, and thus their maxes and means are equivalent. The baseline methods ``RFC", ``Det.", and ``NMF" correspond to the Random Forest Classification, Object Detection AMs, and non-regularized NMF augmented matrix approaches, respectively. Variants of our approach, \textbf{SO}, \textbf{SP}, and \textbf{SOP} correspond to using 
``Spatial+Object Detection" kernels, ``Spatial+Scene Classification" kernels, and ``Spatial+Object Detection+Scene Classification" kernels. Multiple metrics are considered to observe the effects of ground-truth class imbalance, and means are used to quantify performance across a variety of parameter settings.} \label{tab:nsh_as_observations}

\end{table*}

\ifbuildnewimage
\begin{figure*}[ht]
\centering
\footnotesize
\begin{subfigure}[c]{.34\textwidth}
\centering
\includegraphics[height=1.15in]{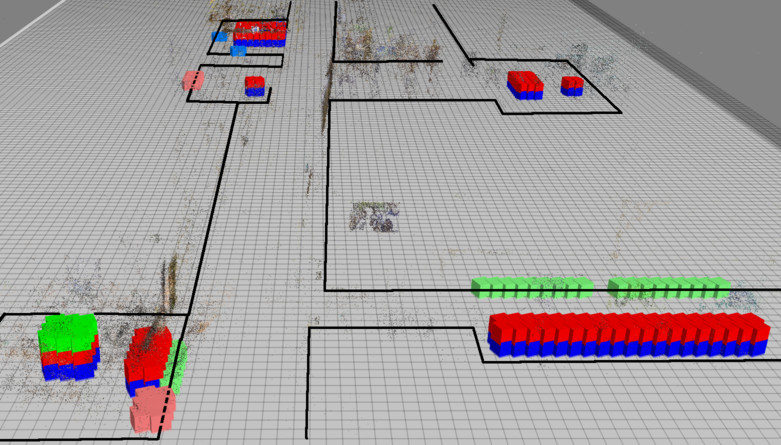}
\caption{\nshscene~GT}
\end{subfigure}
\begin{subfigure}[c]{.30\textwidth}
\centering
\includegraphics[height=1.15in]{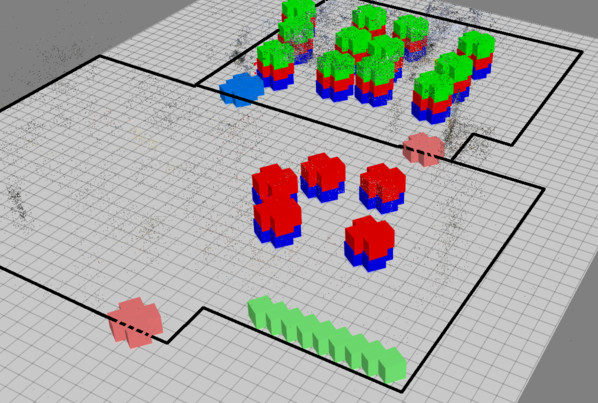}
\caption{\slabscene~GT}
\end{subfigure}
\begin{subfigure}[c]{.20\textwidth}
\centering
\includegraphics[height=1.15in]{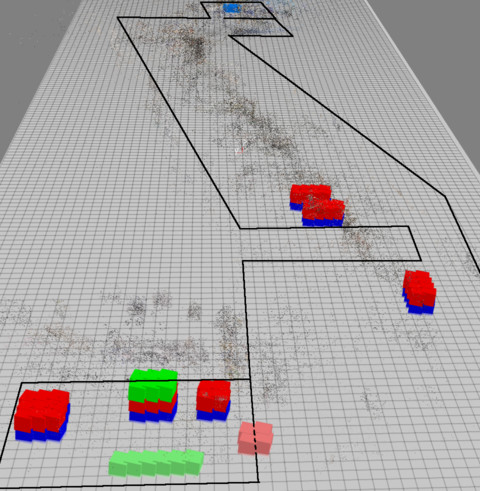}
\caption{\smithscene~GT}
\end{subfigure}
\begin{subfigure}[c]{.25\textwidth}
\centering
\includegraphics[height=1.0in]{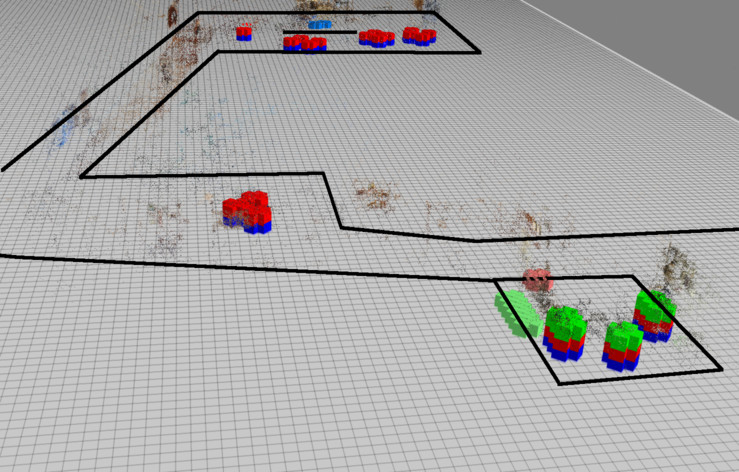}
\caption{\gatesscene~GT}
\end{subfigure}
\begin{subfigure}[c]{.48\textwidth}
\centering
\includegraphics[height=1.0in]{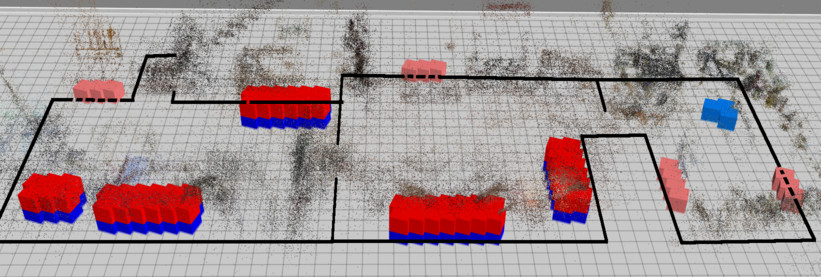}
\caption{\homescene~GT}
\end{subfigure}
\begin{subfigure}[c]{.15\textwidth}
\fcolorbox{black}{white}{\includegraphics[width=\textwidth]{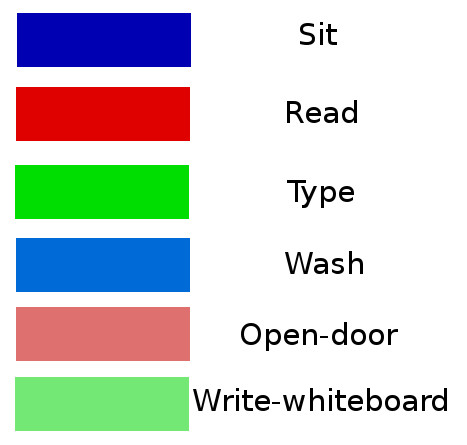}}
\caption{Legend}
\end{subfigure}
\vspace{-7pt}
\caption{\small Ground truth labels and SFM points in each scene. Dotted lines indicate a doorway, solid lines indicate walls.} \label{fig:ground_truth}
\end{figure*}
\fi

Our dataset consists of 5 large, multi-room scenes from various locations. Three scenes, \nshscene, \smithscene, and \gatesscene, are taken from three \textit{distinct} office buildings in the United States, and another scene, \slabscene, comes from an office building in Japan. Each office scene has standard office rooms, common rooms, and a small kitchen area. A final scene, \homescene, consists a kitchen, a living room, and a dining room. See Table~\ref{tab:scene_stats} for scene activity and sparsity statistics. Our goal is to predict dense Action Maps from sparse activity demonstrations.

    The first experiments (Section~\ref{sec:dense-obs-exp}) measure our method's performance when supplied with all observed action data that covers on average about half of all locations and some actions  (See Table~\ref{tab:scene_stats} for the coverage statistics). Additionally, this experiments compares against performance of the spatial kernel-only approach, which serves to illustrate the utility of including side-information.
    However, as it takes some time to collect the observations of each scene, we demonstrate a second set of experiments (Section~\ref{sec:sparse-obs-exp}), to showcase our method handling fractions of the already sparse observation data while still maintaining reasonable performance. In Section~\ref{sec:novel-scene-exp}, our third set of experiments shows that if our method is presented with novel scenes for which there is \textit{zero} activity demonstrations, our method can still make predictions in these new environments. This final set of experiments also investigates which side-information is most helpful for our task.

\subsection{Performance scoring} \label{sec:performance-scoring}

To evaluate an AM, we perform binary classification across all activities and compute mean $F1$ scores. We collect the ground truth activity classes for every image in the scene
by retrieving them from labelled grid cells, as shown in Figure~\ref{fig:ground_truth}, in a small triangle in front of each camera, which represents the viewable space.
We collect the predicted AM scores from the same grid cells and average the scores to produce per-image AM scores.
We used 100 evenly-spaced thresholds to evaluate binary classification performance by averaging $F1$ scores across the thresholds. We report $F1$ scores as opposed to the overall accuracy,
as the overall accuracy of our method is very high due to the 
large amount of space in each scene with no labelled functionality (a large amount of ``true negatives"). The activity classes we use are {\small \texttt{sit, type, open-door, read, write-whiteboard}} and {\small \texttt{wash.}} 
This set of activities provides good coverage of common activities that a person can do in an office or home setting.
To summarize results, we compute the unweighted and 
weighted averages of per-class $F1$ scores, where the weighted average is computed by using the normalized counts of the GT classes in the images. 

\subsection{Preprocessing and parameters} \label{sec:preprocessing}
\ifbuildimage
\begin{figure*}[t]
\centering
\includegraphics[width=.66\textwidth]{images/geometry/room_actual_pic_wgeom_crop_hilite_q90.jpg}
\includegraphics[width=.33\textwidth]{images/room_labels/room_affordance_labels_5_q90.jpg}
   \caption{Left: Picture of small part of labeled floor, and corresponding qualitative geometry automatically generated from the Action Map. Far Right: Labeling tool and labels corresponding to this portion of the scene. Each label indicates that the $(x,y)$ location beneath it has the affordance.}
\label{fig:room_labels}
\end{figure*}
\fi

The first step to build the AM is to build a physical map of the environment. We use Structure-From-Motion (SFM) \cite{wu2013towards} with egocentric videos of a walk through of the environment to obtain a 3D reconstruction of the scene. Next, we consider two important categories of detectable actions: (1) those that involve the user's hands (gesture-based activities), and (2) those that involve significant motion of the user's head, or egomotion-based activities. We used the deep network architecture inspired by \cite{simonyan2014two} to perform activity detection, as the two stream network takes into account both appearance (\eg, hands and objects) as well as motion (\eg, optical flow induced by ego-motion and local hand-object manipulations). When actions are detected by our action recognition module, we need a method for estimating the location of this action. We use the SFM model to compute the 3D camera pose of new images.

As we define an AM over a 2D ground plane (floor layout), we project the 3D camera pose associated to an action to the ground plane. To obtain a ground plane estimate, we fit a plane to a collection of localized cameras using SFM. We assume that the egocentric camera lies approximately at eye level, thus this \textit{height plane} is tangent to the top of the camera wearer's head. We then translate this plane along its normal, while iteratively refitting planes with RANSAC to points in the SFM model. Once we have an estimate of the 2D ground plane in 3D space, we can use it to project the localized actions onto the ground plane. When dealing with multiple scenes, distances must be calibrated between them. We use prior knowledge of the user's height to form estimates of the absolute scale of each scene. Specifically, we use the distance between the ground plane and the user height plane, along with a known user height, to convert distances in the reconstruction to meters. Finally, we grid each scene with cells of size 0.25 meters.  (we use a radius of $2$ grid cells, which is $\sim 0.5$ meters after metric estimation).

Since actions are often strongly correlated with the surrounding area and objects, as shown in Figure~\ref{fig:nsh-feature-examples}, we also extract the visual context of each action as a source of side-information. For every image obtained with the wearable camera, we run scene classification and object detection with \cite{zhou2014learning} and \cite{girshick14CVPR}. We use the pre-trained 
``Places205-GoogLeNet" model for scene-classification, which yields 205 features per image, one per each scene type, and a radius of $2$ grid cells inside which to average the classification scores. For object detection, we use the pretrained ``Bvlc\_reference\_rcnn\_ilsvrc13" model, which performs object detection for 205 different object categories, and use NMS with overlap ratio $0.3$, and min detection score $0.5$.

We use a small grid of parameters for our method ($\alpha \in [0, .1, .3, .5, .7, .9, 1]$, $\lambda \in [10^{-3}, 10^{-2}]$, $\gamma \in [100, 1000]$), where each $\gamma$ is used for the $\chi^2$ kernels, and evaluate performance of multiple 
runs as the cross-run maximum and cross-run average of each of the various scores. In a scenario with
many additional test scenes, a single choice of parameters could be selected via cross-validation.
We also consider variations of our kernel that use different combinations of side-information: Spatial+\objectness~(SO), Spatial+\placeness~(SP), and Spatial+\objectness+\placeness~(SOP). In the first two cases, the $\frac{\alpha}{2}$ weight of Equation~\ref{eq:kernel-combo} becomes $\alpha$ for the \objectness~or \placeness~kernel that is on, and $0$ for the other.

\subsection{Full observation experiments} \label{sec:dense-obs-exp}
When all activity observations are available, our method is able to perform quite well. The dominant source of error is that of camera localization, which reduces the spatial precision of the AM. In Table~\ref{tab:all_scenes_observations}, we evaluate the performance of our method run on each scene separately, as well as running once with all of the scenes in a single matrix. When multiple scenes are used, side-information is crucial: without it, there is no similarity enforced across scenes. In single scene case, we find that using a spatial kernel only can perform well, yet is generally outperformed by using all side information, especially when side information and activity demonstrations are present from other scenes. By using the data from all scenes simultaneously in a global factorization, performance increases globally over using each single scene's data alone. This is expected and desirable: simultaneous understanding of multiple scenes can improve as the set of available scenes with observation data grows.
 
\subsection{Partial observation experiments} \label{sec:sparse-obs-exp}
 We expose our algorithm to various fractions of the total activity demonstrations to simulate an increasing amount of observed actions. We find that performance is high even with only a few demonstrations and steadily increases as the amount of activity demonstrations increases.  The \nshscene, \smithscene, and \homescene~scenes have enough activity demonstration data to illustrate the performance gains
of our method as a function of the 
available data. We show quantitative per-class results for these in Figure~\ref{fig:elapse-graphs}. Sharp increases can be observed in the per-class trends, which correspond to the increase of coverage of each activity class. In Figure~\ref{fig:elapse-overhead1}, we
show the overhead view of the AM for the {\small \texttt{sit}} and {\small \texttt{type}} labels for the \nshscene~as a function of the available data, where it can be seen how the AM qualitatively improves over time as observations are collected.

\subsection{Novel scene experiments} \label{sec:novel-scene-exp}

Another scenario is the task of predicting AMs for novel scenes containing \textit{zero} activity observation data. Our method leverages the appearance and activity observation data in one scene, and only appearance data in the novel scene to make predictions. We now introduce three baselines we consider. 
The first baseline is to perform per-image classification with the \objectness~and \placeness~features,
which serves to estimate image-wise performance of using the \objectness~and \placeness~information. This baseline requires observations in a labelled scene for training. We use Random Forests \cite{breiman2001random} as the classification method, trained on images from the source scene.
The second baseline we consider is non-regularized Weighted Nonnegative Matrix Factorization by augmenting the target 
matrix $\mathbf{R}$ with the \objectness~and \placeness~features for each location.
This baseline does not explicitly enforce the similarity that the regularized framework does, thus, we expect it
to not perform as well as our framework. 
The third baseline we consider is to build AMs from the back-projected object detections by directly
associating each detection category with an activity category. 

We use the \nshscene~demonstration and appearance data as input and evaluate the performance
by applying the learned model to each of the other scenes. These results (Table~\ref{tab:nsh_as_observations}) illustrate that our method's AM predictions outperform the baselines in $\frac{13}{16}$ cases, and that the appearance information is capitalized
upon the most by our method.
We find that \placeness~is particularly beneficial to performance, a phenomenon for which we present two hypothesized factors: 1) as shown in \cite{zhou2014object} ``object detectors emerge in deep scene CNNs", suggesting that the Scene Classification features subsume the cues present in the object detector
features, and 2) due to localization noise, correlations between localized activities and localized objects are 
not as strong, and can serve to introduce noise to the Spatial+Scene Classification kernel combination when this object information is integrated. 

Overall, we find that our model harnesses the power of activity observations in concert with the availability of rich scene classification and object detection information to estimate the functionality of environments both with and without activity observations. See Appendix~\ref{sec:appendix_examples}, including Tables~\ref{fig:full-example1}~and~\ref{fig:full-example2} for additional visualizations and novel scene prediction demonstrations.

\ifbuildnewimage
\newlength{\elapsefiglen}
\setlength{\elapsefiglen}{.14\textwidth}

\begin{figure}[t]
\centering
\footnotesize
\begin{subfigure}[c]{\elapsefiglen}
\includegraphics[angle=90,width=\textwidth]{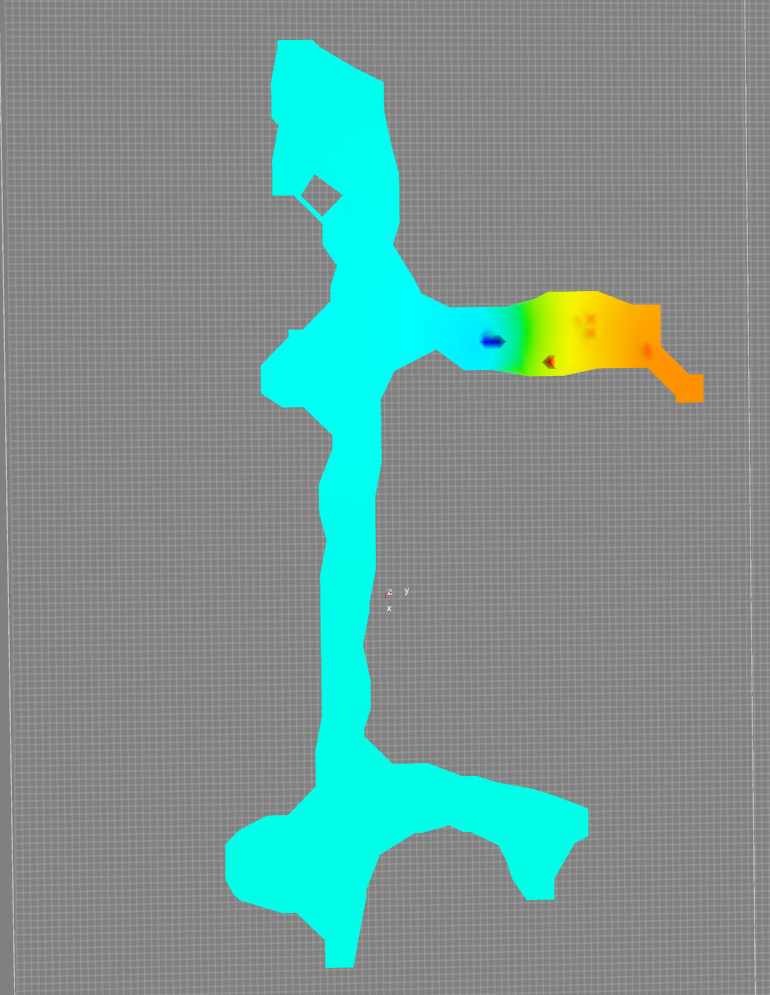}
\includegraphics[angle=90,width=\textwidth]{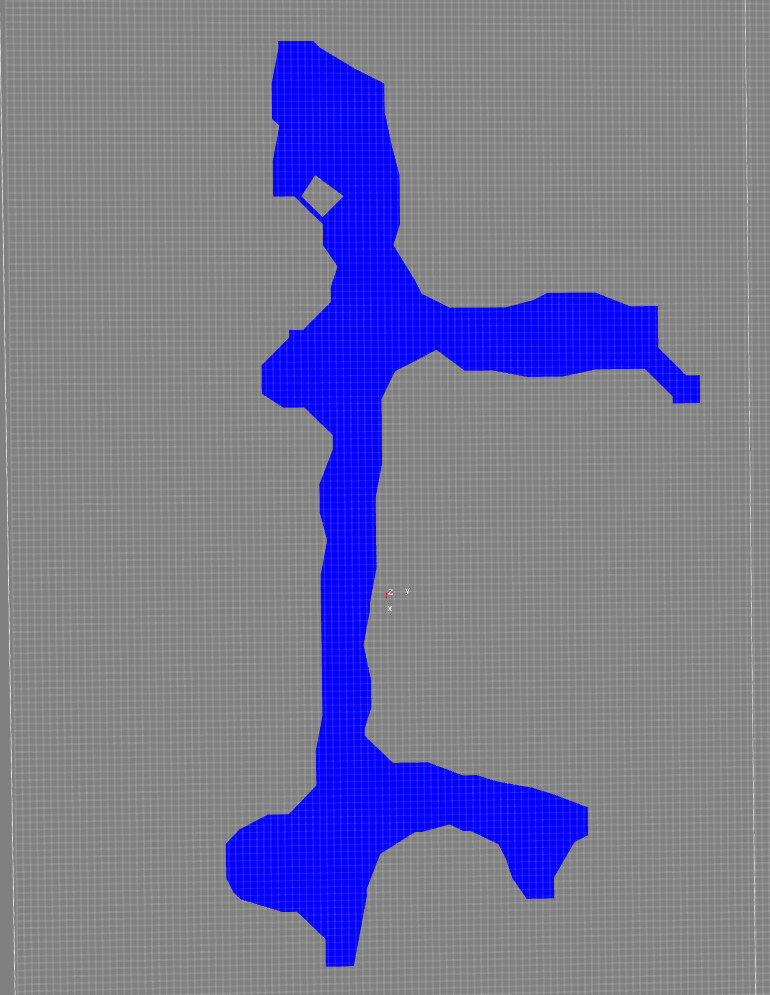}
\end{subfigure}
\begin{subfigure}[c]{\elapsefiglen}
\includegraphics[angle=90,width=\textwidth]{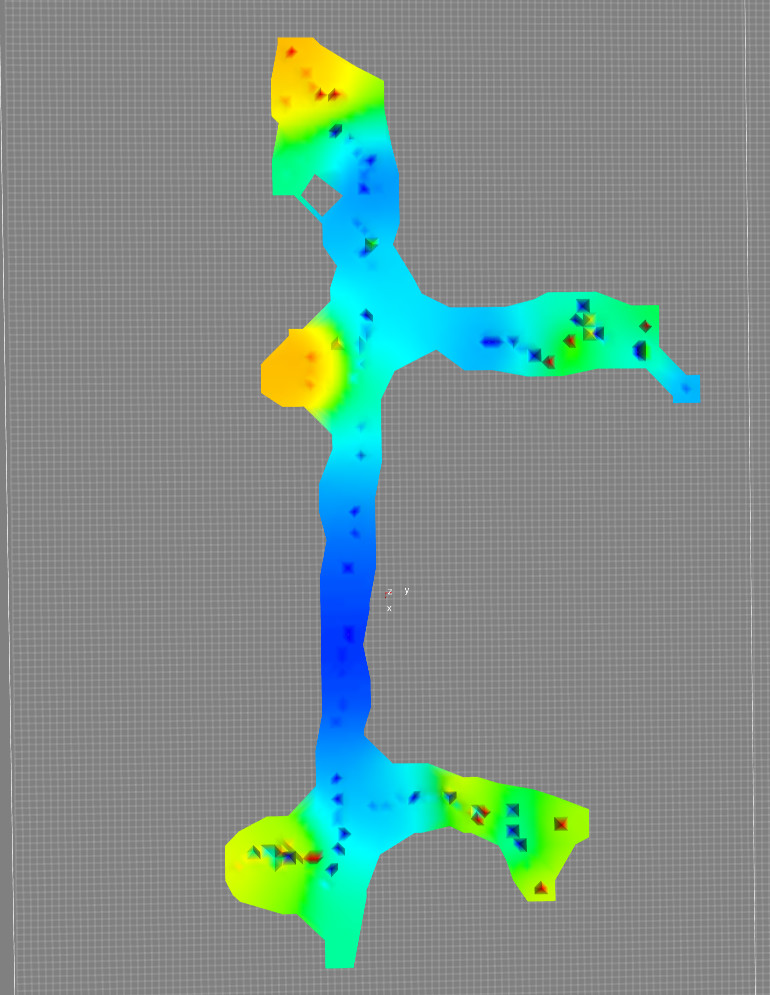}
\includegraphics[angle=90,width=\textwidth]{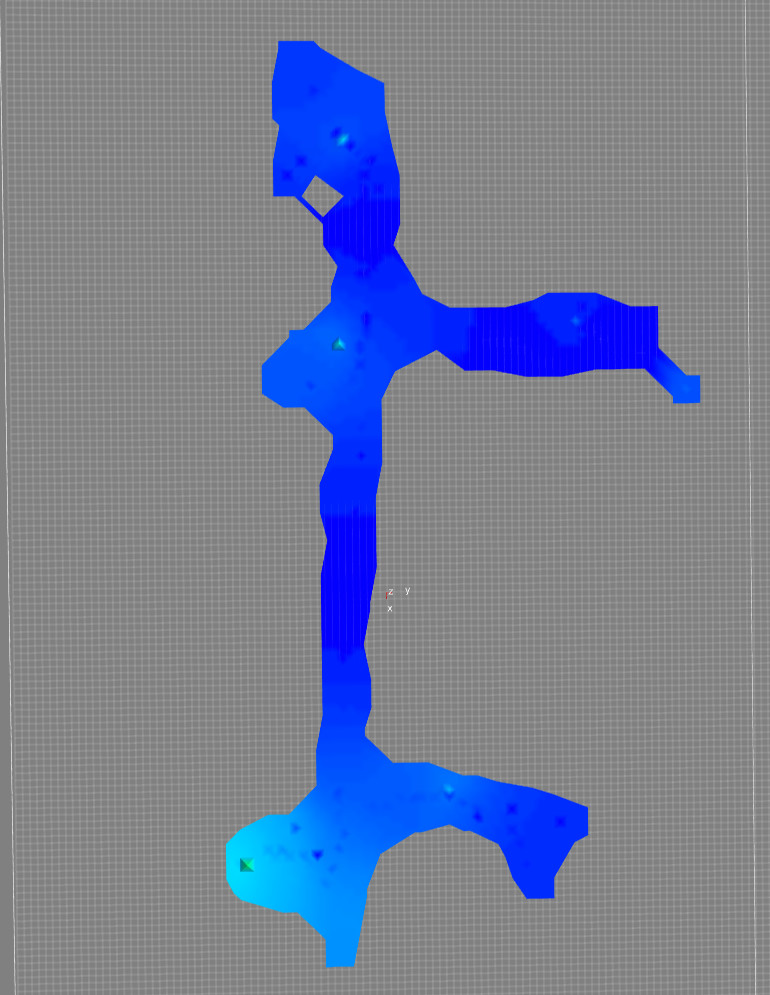}
\end{subfigure}
\begin{subfigure}[c]{\elapsefiglen}
\includegraphics[angle=90,width=\textwidth]{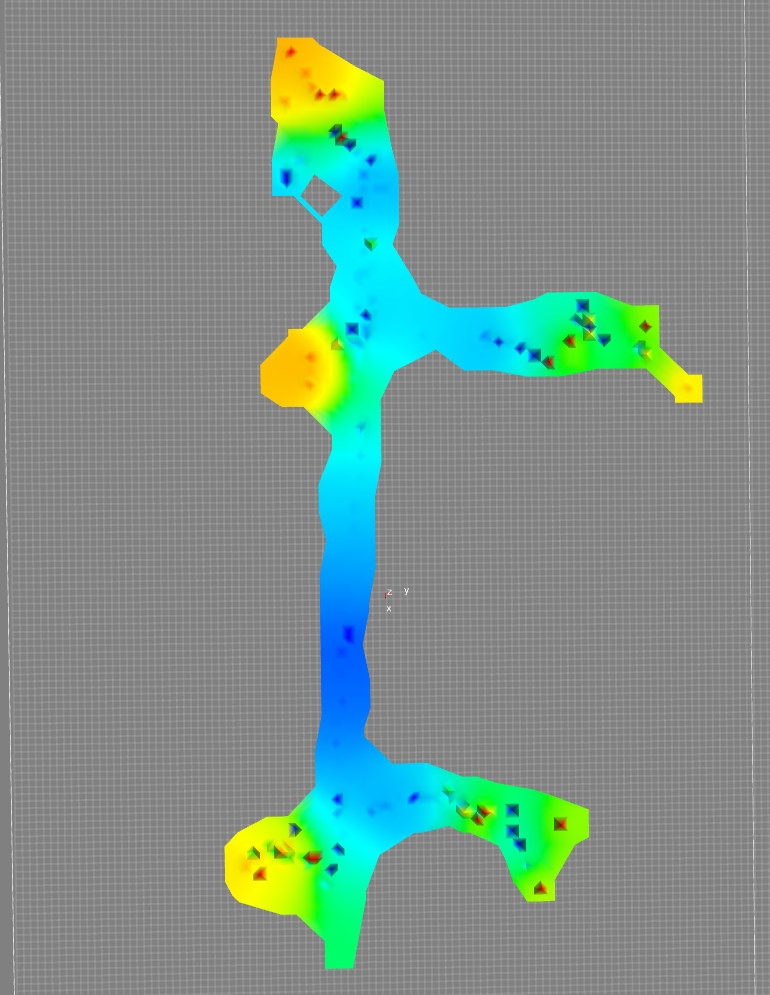}
\includegraphics[angle=90,width=\textwidth]{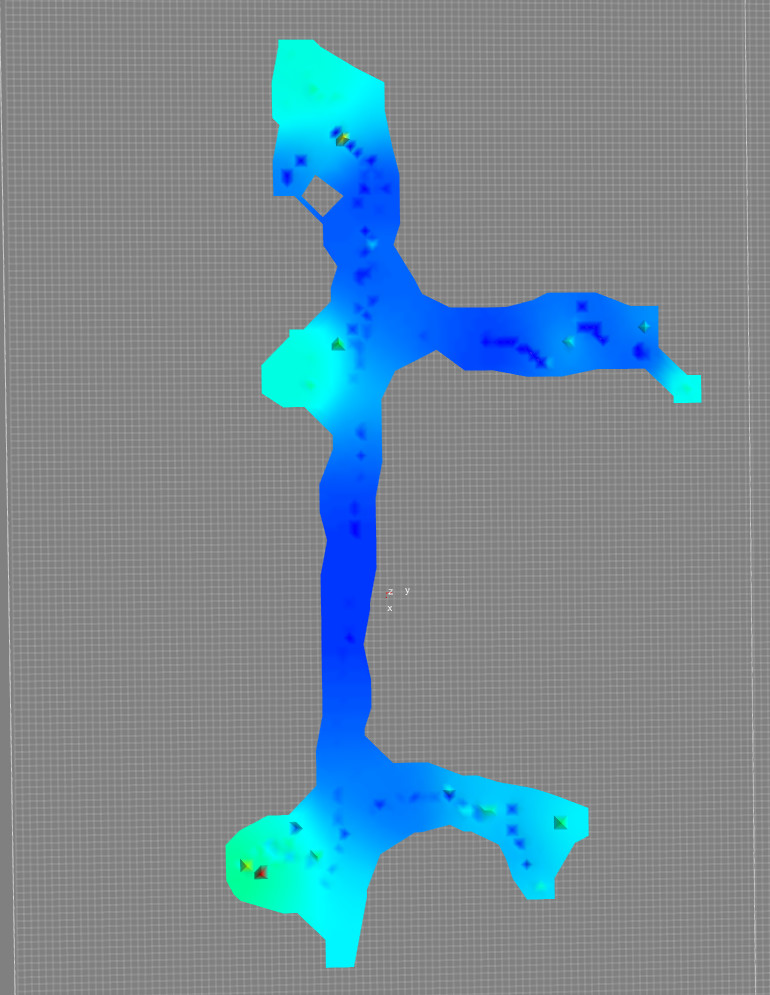}
\end{subfigure}
\vspace{-7pt}
  \caption{\small 'Sit' (top row) and 'Type' (bottom row) AMs as the amount of observed data increases on \nshscene. The columns stand for $10\%$, $80\%$, and $100\%$ of the data.}
\label{fig:elapse-overhead1}
\end{figure}
\fi

\section{Action Maps for Localization} \label{sec:applications}

\ifbuildnewimage
\begin{figure}[t]
\centering
\footnotesize
\includegraphics[width=.8\columnwidth]{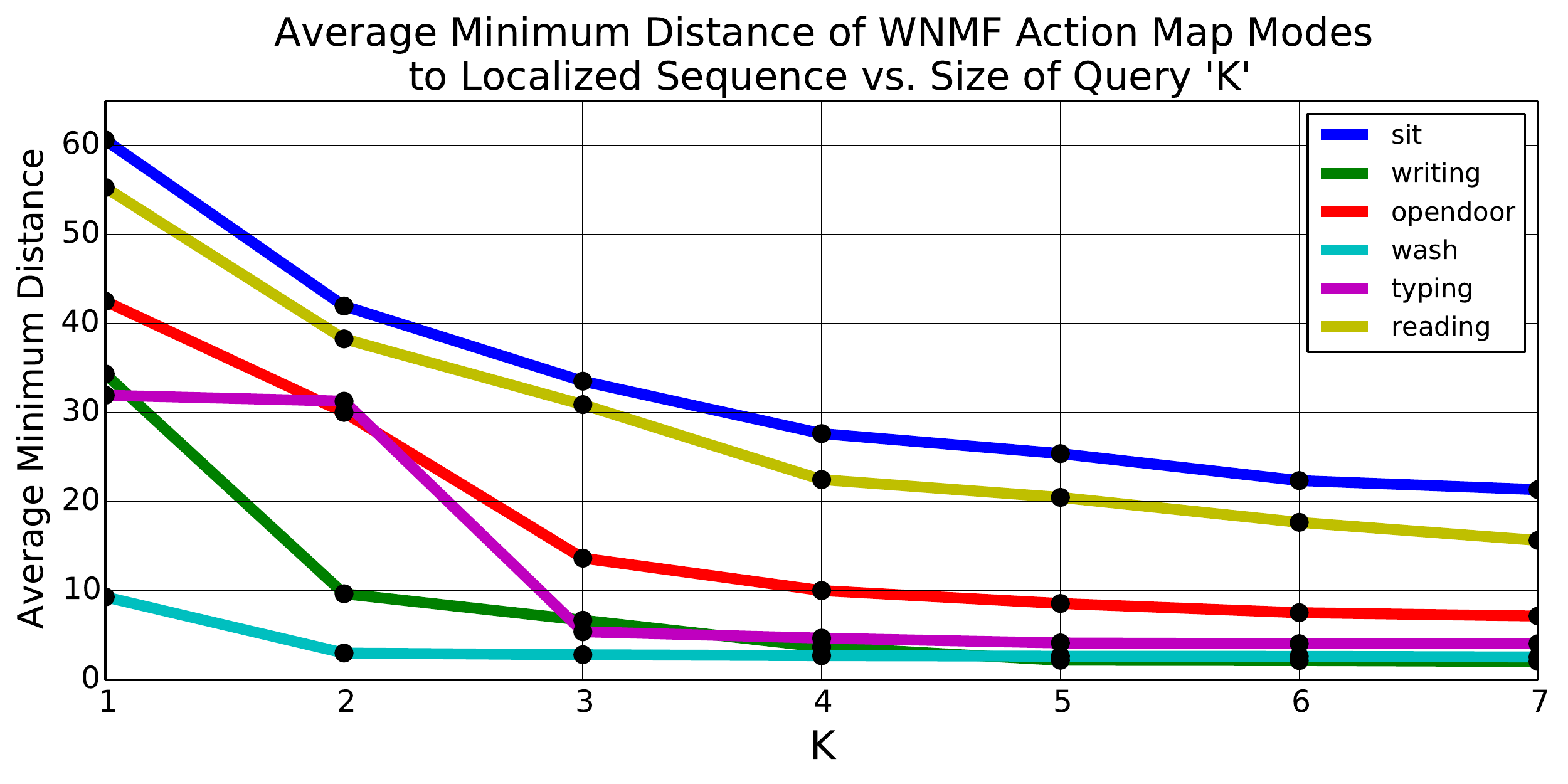}
\vspace{-7pt}
\caption{\small Localizing with an Action Map and observed activities. Activities that are more specialized are localized with less guesses.} \label{fig:am-localize-plussing}
\end{figure}
\fi

We demonstrate a proof-of-concept application of Action Maps to the task of localization.  Intuitively, by leveraging the ``where an activity can be done" functional-spatial information from Action Maps, along with ``what activity has been done" functional information from activity detection, the user's spatial location is constrained to be in one of several areas. We localize activity sequences in each 2D map based on the combination of predicted action locations from the Action Map, and observed actions in each frame. In Figure~\ref{fig:am-localize-plussing}, we show the spatial discrepancy in grid cells between the $K$-best AM location guesses decreases. Thus, an Action Map can be used to localize a person with observations of their activity.

\newcommand*{\vcenteredhbox}[1]{\begingroup
\setbox0=\hbox{#1}\parbox{\wd0}{\box0}\endgroup}
\newlength{\pflength}
\setlength{\pflength}{.25\textwidth}

\section{Conclusion}

We have demonstrated a novel method for generating functional maps of uninstrumented common environments. Our model jointly considers scene appearance and functionality while consolidating evidence from the natural vantage point of the user, and is able to learn from a user's demonstrations to make predictions of functionality of less explored and completely novel areas. Finally, our proof-of-concept application hints at the breadth of future work that can exploit the rich spatial and functional information present in Action Maps. 

\section*{Acknowledgements}
This research was funded in part by grants from the PA Dept. of Health's Commonwealth Universal Research Enhancement Program, IBM Research Open Collaborative Research initiative, CREST (JST), and an NVIDIA hardware grant. We thank Ryo Yonetani for valuable data collection assistance and discussion.

{\small
\bibliographystyle{ieee}
\bibliography{egbib}

\begin{thebibliography}{10}\itemsep=-1pt

\bibitem{breiman2001random}
L.~Breiman.
\newblock Random forests.
\newblock {\em Machine learning}, 45(1):5--32, 2001.

\bibitem{delaitre2012scene}
V.~Delaitre, D.~F. Fouhey, I.~Laptev, J.~Sivic, A.~Gupta, and A.~A. Efros.
\newblock Scene semantics from long-term observation of people.
\newblock In {\em Computer Vision--ECCV 2012}, pages 284--298. Springer, 2012.

\bibitem{fathi2011understanding}
A.~Fathi, A.~Farhadi, and J.~M. Rehg.
\newblock Understanding egocentric activities.
\newblock In {\em Computer Vision (ICCV), 2011 IEEE International Conference
  on}, pages 407--414. IEEE, 2011.

\bibitem{Fouhey12}
D.~F. Fouhey, V.~Delaitre, A.~Gupta, A.~A. Efros, I.~Laptev, and J.~Sivic.
\newblock People watching: Human actions as a cue for single-view geometry.
\newblock In {\em Proc. 12th European Conference on Computer Vision}, 2012.

\bibitem{gall2011functional}
J.~Gall, A.~Fossati, and L.~Van~Gool.
\newblock Functional categorization of objects using real-time markerless
  motion capture.
\newblock In {\em Computer Vision and Pattern Recognition (CVPR), 2011 IEEE
  Conference on}, pages 1969--1976. IEEE, 2011.

\bibitem{girshick14CVPR}
R.~Girshick, J.~Donahue, T.~Darrell, and J.~Malik.
\newblock Rich feature hierarchies for accurate object detection and semantic
  segmentation.
\newblock In {\em Computer Vision and Pattern Recognition}, 2014.

\bibitem{gu2010collaborative}
Q.~Gu, J.~Zhou, and C.~H. Ding.
\newblock Collaborative filtering: Weighted nonnegative matrix factorization
  incorporating user and item graphs.
\newblock In {\em SDM}, pages 199--210. SIAM, 2010.

\bibitem{gupta2008context}
A.~Gupta, T.~Chen, F.~Chen, D.~Kimber, and L.~S. Davis.
\newblock Context and observation driven latent variable model for human pose
  estimation.
\newblock In {\em Computer Vision and Pattern Recognition, 2008. CVPR 2008.
  IEEE Conference on}, pages 1--8. IEEE, 2008.

\bibitem{GuptaSatkinEfrosHebert_CVPR11}
A.~Gupta, S.~Satkin, A.~A. Efros, and M.~Hebert.
\newblock From 3d scene geometry to human workspace.
\newblock In {\em Computer Vision and Pattern Recognition(CVPR)}, 2011.

\bibitem{jiang2013hallucinated}
Y.~Jiang, H.~Koppula, and A.~Saxena.
\newblock Hallucinated humans as the hidden context for labeling 3d scenes.
\newblock In {\em Computer Vision and Pattern Recognition (CVPR), 2013 IEEE
  Conference on}, pages 2993--3000. IEEE, 2013.

\bibitem{koppula2013learning}
H.~S. Koppula, R.~Gupta, and A.~Saxena.
\newblock Learning human activities and object affordances from rgb-d videos.
\newblock {\em The International Journal of Robotics Research}, 32(8):951--970,
  2013.

\bibitem{li2015delving}
Y.~Li, Z.~Ye, and J.~M. Rehg.
\newblock Delving into egocentric actions.
\newblock In {\em Proceedings of the IEEE Conference on Computer Vision and
  Pattern Recognition}, pages 287--295, 2015.

\bibitem{moore1999exploiting}
D.~J. Moore, I.~Essa, M.~H. Hayes~III, et~al.
\newblock Exploiting human actions and object context for recognition tasks.
\newblock In {\em Computer Vision, 1999. The Proceedings of the Seventh IEEE
  International Conference on}, volume~1, pages 80--86. IEEE, 1999.

\bibitem{SilbermanECCV12}
P.~K. Nathan~Silberman, Derek~Hoiem and R.~Fergus.
\newblock Indoor segmentation and support inference from rgbd images.
\newblock In {\em ECCV}, 2012.

\bibitem{peursum2005combining}
P.~Peursum, G.~West, and S.~Venkatesh.
\newblock Combining image regions and human activity for indirect object
  recognition in indoor wide-angle views.
\newblock In {\em Computer Vision, 2005. ICCV 2005. Tenth IEEE International
  Conference on}, volume~1, pages 82--89. IEEE, 2005.

\bibitem{pirsiavash2012detecting}
H.~Pirsiavash and D.~Ramanan.
\newblock Detecting activities of daily living in first-person camera views.
\newblock In {\em Computer Vision and Pattern Recognition (CVPR), 2012 IEEE
  Conference on}, pages 2847--2854. IEEE, 2012.

\bibitem{ramachandran2011mayavi}
P.~Ramachandran and G.~Varoquaux.
\newblock {Mayavi: 3D Visualization of Scientific Data}.
\newblock {\em Computing in Science \& Engineering}, 13(2):40--51, 2011.

\bibitem{savva2014scenegrok}
M.~Savva, A.~X. Chang, P.~Hanrahan, M.~Fisher, and M.~Nie{\ss}ner.
\newblock Scenegrok: Inferring action maps in 3d environments.
\newblock {\em ACM Transactions on Graphics (TOG)}, 33(6), 2014.

\bibitem{simonyan2014two}
K.~Simonyan and A.~Zisserman.
\newblock Two-stream convolutional networks for action recognition in videos.
\newblock In {\em Advances in Neural Information Processing Systems}, pages
  568--576, 2014.

\bibitem{De_la_Torre_Frade_2009_6394}
E.~H. Spriggs, F.~{De la Torre Frade}, and M.~Hebert.
\newblock Temporal segmentation and activity classification from first-person
  sensing.
\newblock In {\em IEEE Workshop on Egocentric Vision, CVPR 2009}, June 2009.

\bibitem{wu2013towards}
C.~Wu.
\newblock Towards linear-time incremental structure from motion.
\newblock In {\em 3D Vision-3DV 2013, 2013 International Conference on}, pages
  127--134. IEEE, 2013.

\bibitem{yao2010modeling}
B.~Yao and L.~Fei-Fei.
\newblock Modeling mutual context of object and human pose in human-object
  interaction activities.
\newblock In {\em Computer Vision and Pattern Recognition (CVPR), 2010 IEEE
  Conference on}, pages 17--24. IEEE, 2010.

\bibitem{zhou2014object}
B.~Zhou, A.~Khosla, A.~Lapedriza, A.~Oliva, and A.~Torralba.
\newblock Object detectors emerge in deep scene cnns.
\newblock {\em arXiv preprint arXiv:1412.6856}, 2014.

\bibitem{zhou2014learning}
B.~Zhou, A.~Lapedriza, J.~Xiao, A.~Torralba, and A.~Oliva.
\newblock Learning deep features for scene recognition using places database.
\newblock In {\em Advances in Neural Information Processing Systems}, pages
  487--495, 2014.

\end{thebibliography}
}

\appendix
\clearpage
\section{Examples} \label{sec:appendix_examples}
In the following examples displayed in Tables~\ref{fig:full-example1}~and~\ref{fig:full-example2}, we display visualizations of various components of our method. In both examples, activity demonstration data is available for only one scene. In Table~\ref{fig:full-example1}, we display object detection results for images from each scene, as well as projected onto the floor planes. In Table~\ref{fig:full-example2}, we display scene classification results for images from each scene, as well as projected onto the floor planes. In Table~\ref{fig:full-example1}, the target scene without activity demonstration is a scene from the NYU V2 Depth Dataset \cite{SilbermanECCV12} reconstructed and processed by our method. The per-row verbose description is as follows.\footnote{We used \cite{ramachandran2011mayavi} to produce 3D visualizations throughout the paper.}

\begin{enumerate}
\item Scene Names
\item Scene reconstruction with localized cameras visualized as vectors, colored by temporal ordering
\item (Object detections) or (Scene classifications), with scores for example images \label{item:obj}
\item (Object detections and \texttt{sit})
 or (Scene classification \texttt{corridor}) features visualized, with the example images (and objects from row \ref{item:obj}) localized in the scene
\item Available localized activity demonstrations, height corresponds to bin count, color corresponds to activity type (with the same coloration scheme as Figure~\ref{fig:ground_truth})
\item Final \texttt{sit} Action Maps as produced by our method visualized in 3D
\item Final \texttt{sit} Action Maps as produced by our method visualized as projected onto the example images (with no occlusion filtering).
\end{enumerate}

\begin{table*}[t]
\centering
\footnotesize
%% \begin{tabularx}{\textwidth}{|m{2cm}|X|X|}
\begin{tabularx}{\textwidth}{@{}|m{2cm}|Y|Y|@{}}
\hline
  & \homescene (training) & NYUV2 Home Office 0001 \\
\hline
 Scene Reconstructions with Localized Cameras&
\includegraphics[width=0.35\textwidth,trim={0cm 0cm 2cm 0cm},clip=true]{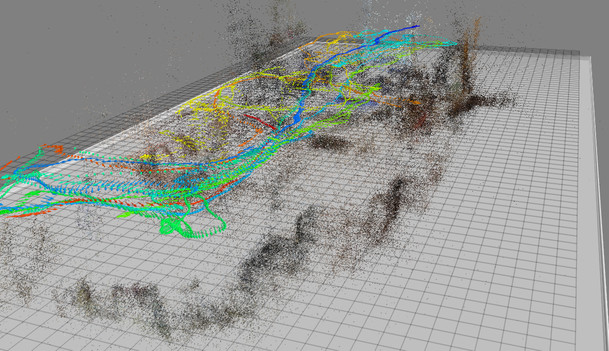} &
\includegraphics[width=.35\textwidth,trim={0cm 0cm 2cm 3cm},clip=true]{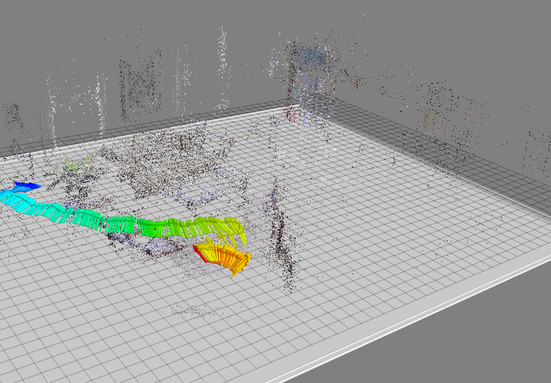} \\
\hline
Example Images with Detections &
\includegraphics[width=.35\textwidth,trim={10cm 1cm 10cm 1cm},clip=true]{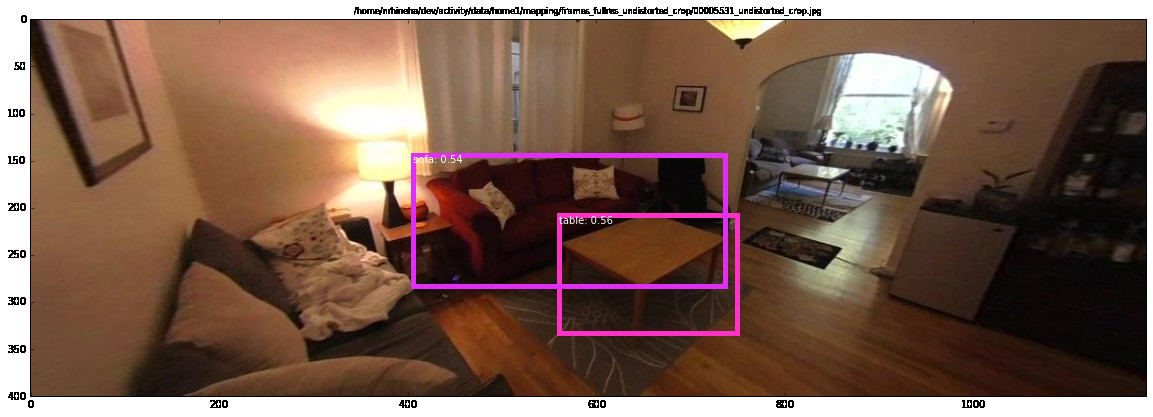} &
\includegraphics[width=.35\textwidth,trim={1cm 1cm 1cm 3cm},clip=true]{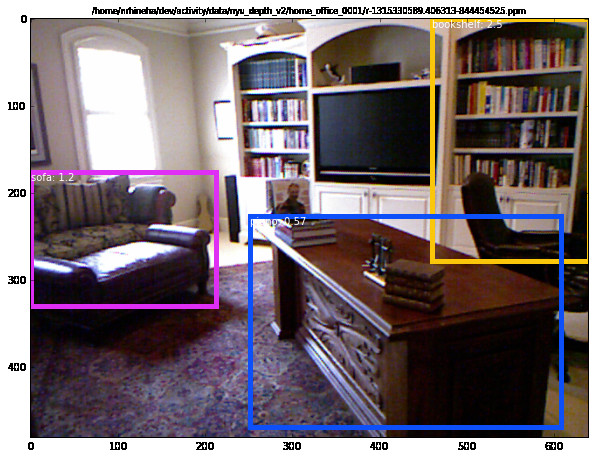} \\
\hline
Object Detections in 3D and Object `\texttt{sit}' Action Maps &
\includegraphics[width=.35\textwidth,trim={0cm 0cm 2cm 0cm},clip=true]{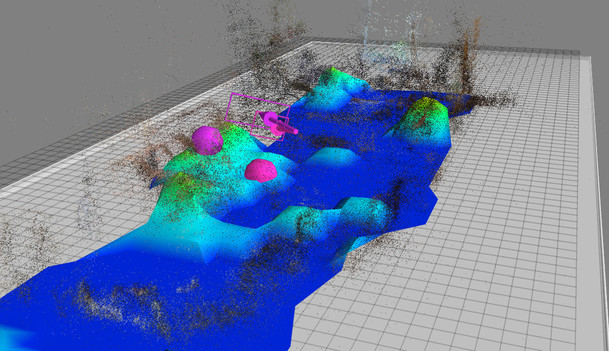} &
\includegraphics[width=.33\textwidth]{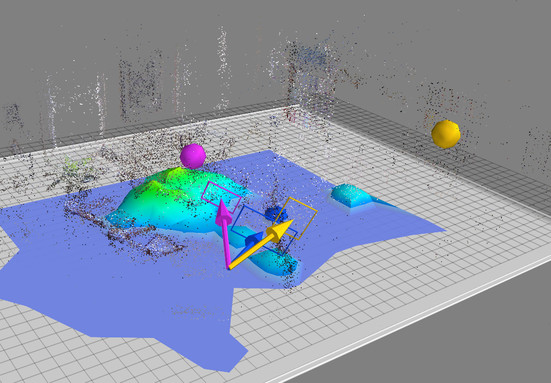} \\
\hline
Binned Localized Detected Actions (colored by type) %% \fcolorbox{black}{white}{\includegraphics[width=3.5cm]{new_images/gt_ims/legend_q90.jpg}} 
%% (Legend above)
&
\includegraphics[width=.35\textwidth,trim={0cm 0cm 5cm 0cm},clip=true]{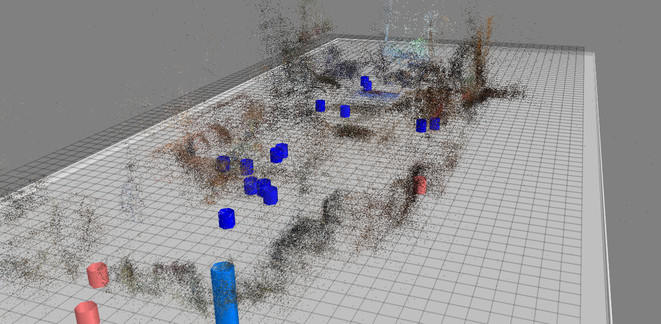} &
None - Novel Scene \\
\hline
`\texttt{sit}' Action Maps &
\includegraphics[width=.35\textwidth,trim={0cm 0cm 5cm 0cm},clip=true]{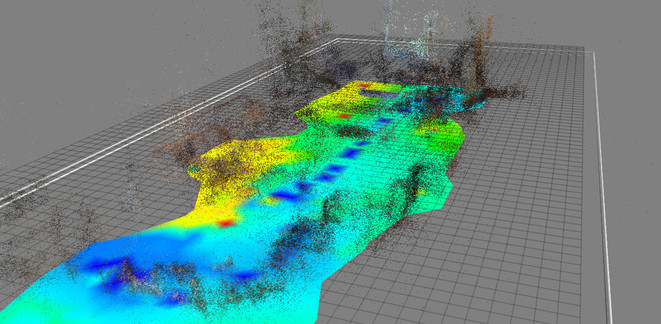} &
\includegraphics[width=.35\textwidth,trim={0cm 0cm 0cm 0cm},clip=true]{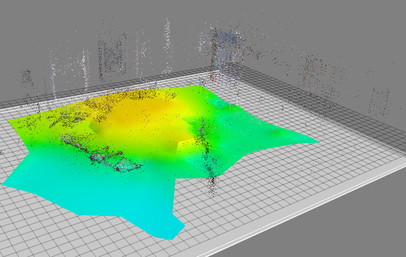} \\
\hline
Projected `\texttt{sit}' Action Maps &
\includegraphics[width=.35\textwidth,trim={10cm 1cm 10cm 1cm},clip=true]{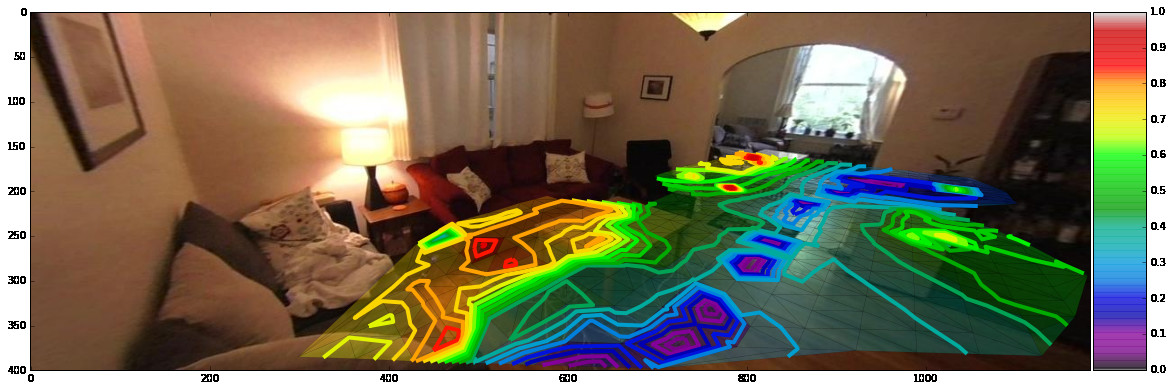} &
\includegraphics[width=.35\textwidth,trim={1cm 1cm 1cm 3cm},clip=true]{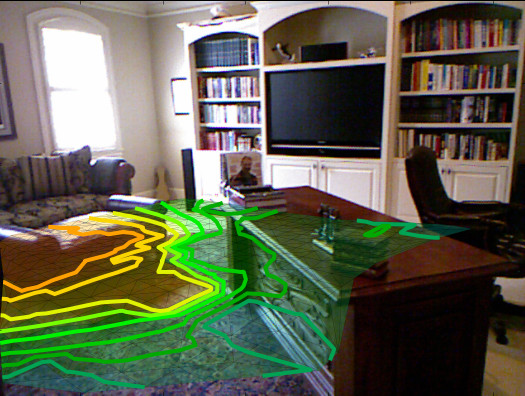} \\
\hline

\end{tabularx}
\caption{\small Visualizations of various aspects of the method.}
\label{fig:full-example1}
\end{table*}

\begin{table*}[t]
\centering
\footnotesize
%% \begin{tabularx}{\textwidth}{|m{2.5cm}|X|p{.44\textwidth}|}
\begin{tabularx}{\textwidth}{@{}|m{2.5cm}|Y|L{.44\textwidth}|@{}}
\hline
& \nshscene (training) & \slabscene \\
\hline
 Scene Reconstructions with Localized Cameras&
\includegraphics[width=0.35\textwidth,trim={0cm 0cm 4cm 0cm},clip=true]{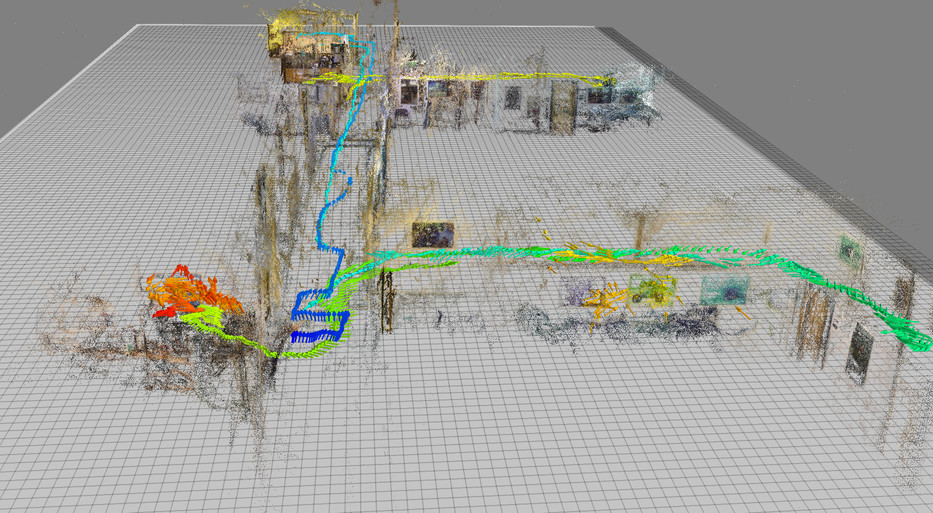} &
\noindent\parbox[c]{\hsize}{\includegraphics[width=.44\textwidth,trim={0cm 0cm 0cm 0cm},clip=true]{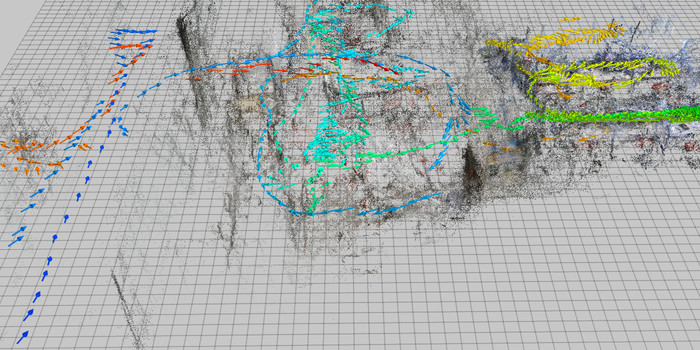}} \\
\hline
Example Images with Classification Score&
\includegraphics[width=.35\textwidth,%% trim={10cm 1cm 10cm 1cm},clip=true
]{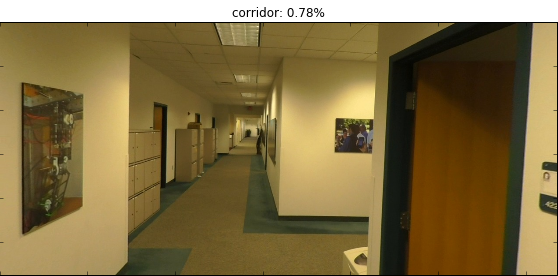} &
\noindent\parbox[c]{\hsize}{\includegraphics[width=.44\textwidth,trim={0cm 1.5cm 1.5cm 0cm},clip=true
]{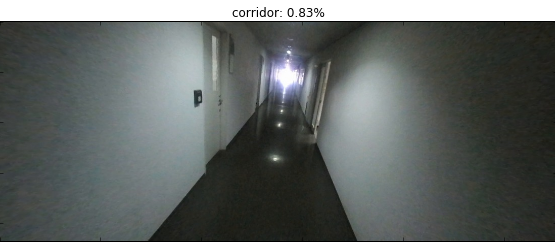}} \\
\hline
\texttt{Corridor} Scene Classifications in 3D with Localized Example Images&
\includegraphics[width=.35\textwidth,trim={0cm 0cm 5cm 0cm},clip=true]{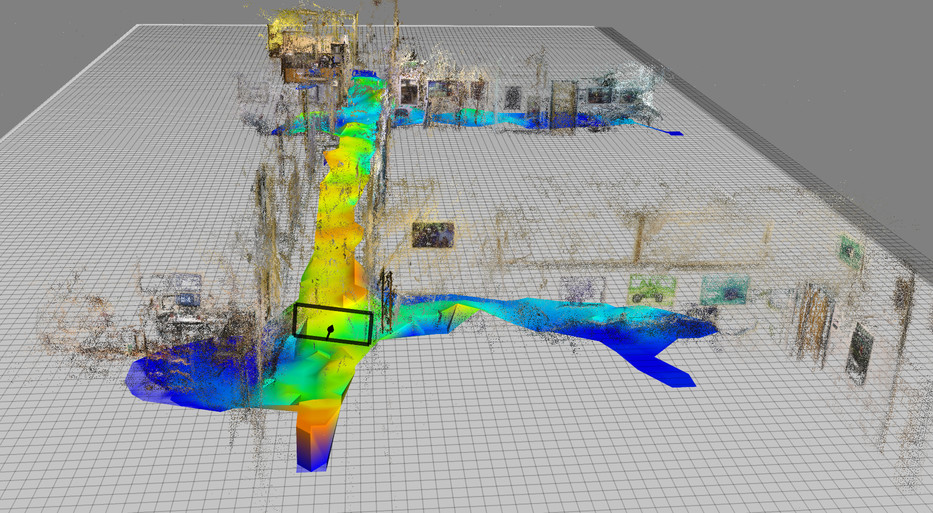} &
\noindent\parbox[c]{\hsize}{\includegraphics[width=.44\textwidth,trim={0cm 0cm 0cm 0cm},clip=true]{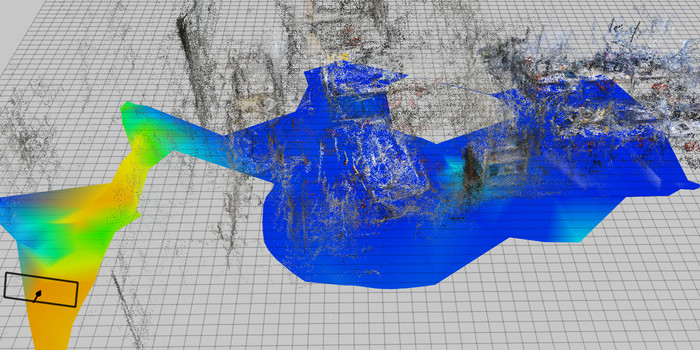}} \\
\hline
Binned Localized Detected Actions (colored by type) %% \fcolorbox{black}{white}{\includegraphics[width=3.5cm]{new_images/gt_ims/legend_q90.jpg}} 
%% (Legend above)
&
\includegraphics[width=.35\textwidth,trim={0cm 0cm 5cm 0cm},clip=true]{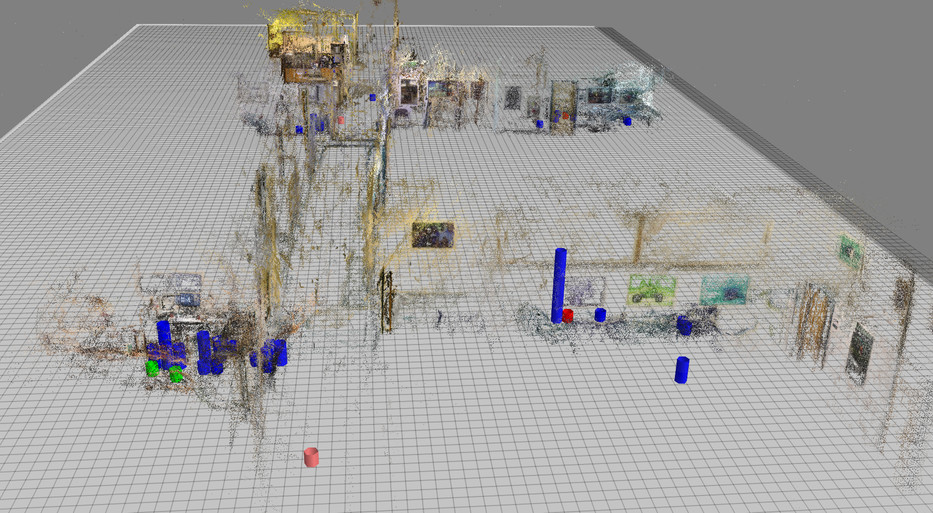} &
None - Novel Scene \\
\hline
`\texttt{sit}' Action Maps &
\includegraphics[width=.35\textwidth,trim={0cm 0cm 2.5cm 0cm},clip=true]{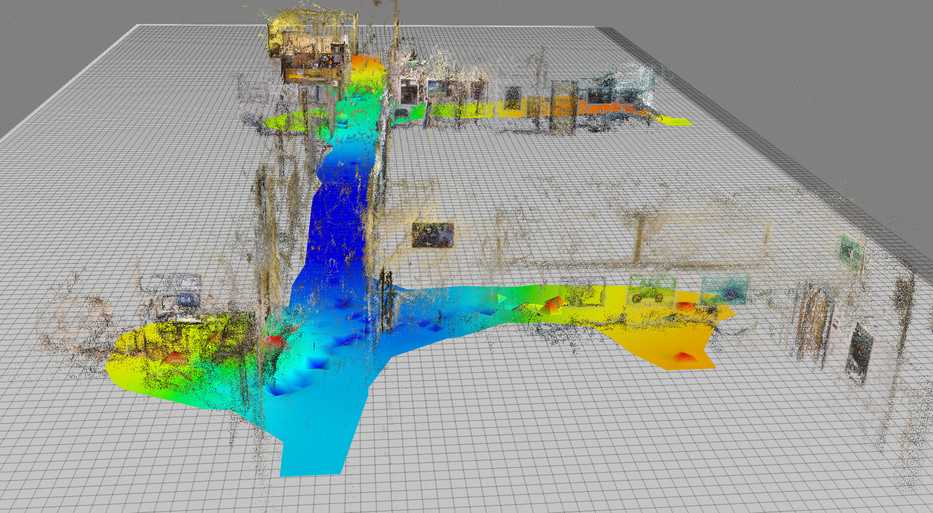} &
\noindent\parbox[c]{\hsize}{\includegraphics[width=.44\textwidth,trim={0cm 2cm 0cm 0cm},clip=true]{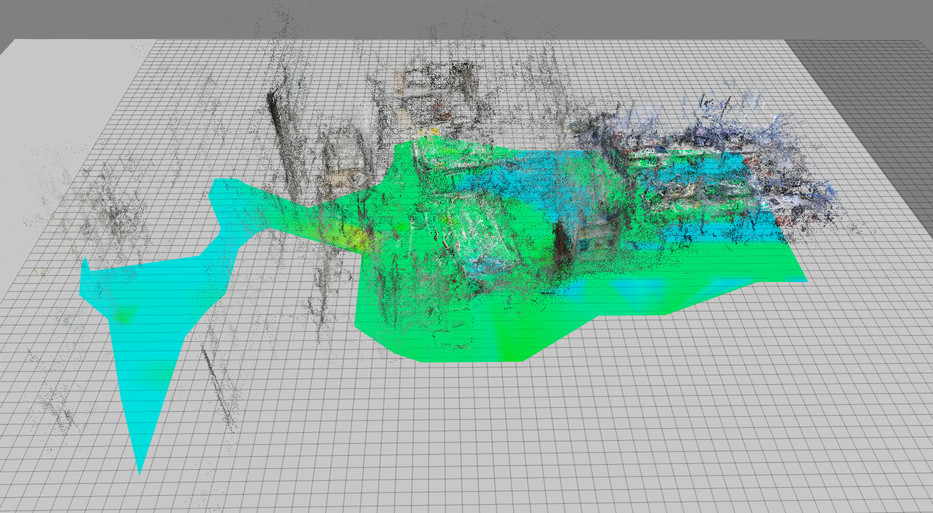}} \\
\hline
Projected `\texttt{sit}' Action Maps &
\includegraphics[width=.35\textwidth%% ,trim={10cm 1cm 10cm 1cm},clip=true
]{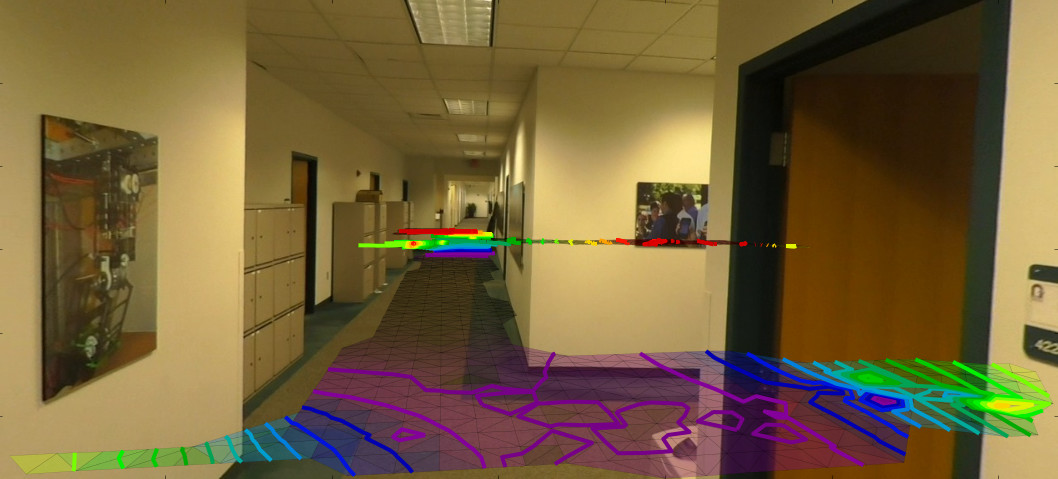} &
\noindent\parbox[c]{\hsize}{\includegraphics[width=.44\textwidth,trim={0cm 1cm 1.5cm 0cm},clip=true]{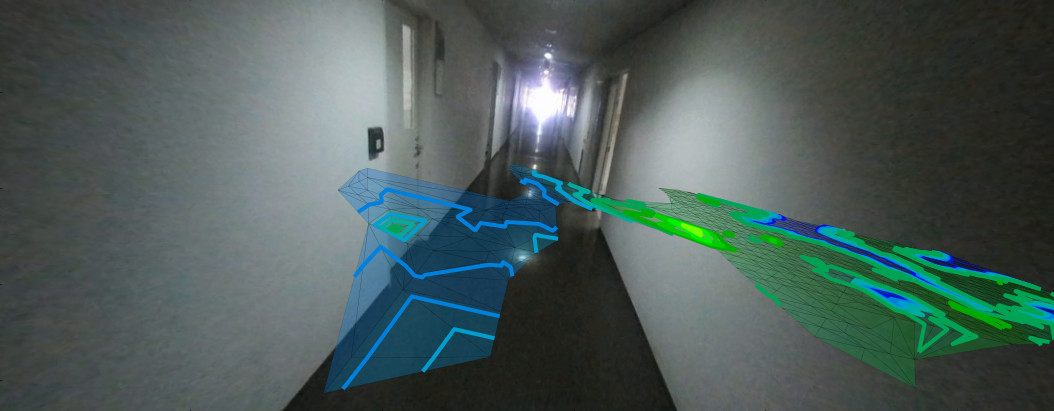}} \\
\hline

\end{tabularx}
\caption{\small Visualizations of various aspects of the method}
\label{fig:full-example2}
\end{table*}
\end{document}